 \documentclass[pmlr,twocolumn,10pt]{jmlr} % W&CP article

\usepackage[switch]{lineno}

\newcommand{\equal}[1]{{\hypersetup{linkcolor=black}\thanks{#1}}}

\theorembodyfont{\upshape}
\theoremheaderfont{\scshape}
\theorempostheader{:}
\theoremsep{\newline}

\usepackage{my}
\usepackage{misc}
\usepackage{mytikz}
\usepackage{microtype}
\usepackage{booktabs}

\jmlrvolume{297}
\jmlryear{2025}
\jmlrworkshop{Machine Learning for Health (ML4H) 2025} % W&CP title

 \title[KarmaTS]{KarmaTS: A Universal Simulation Platform for Multivariate Time Series with Functional Causal Dynamics}

\author{
\Name{Haixin Li}\equal{These authors contributed equally} 
\Email{peter.li@tum.de}\\ 
\addr School of Computation, Information and Technology (CIT)\\
\addr Technical University of Munich, Germany \\
\addr SCAI Lab, D-HEST, ETH Zurich, Switzerland
\AND 
\Name{Yanke Li}\footnotemark[1]
\Email{yanke.li@hest.ethz.ch}\\
\addr SCAI Lab, D-HEST, ETH Zurich, Switzerland \\ 
\addr Digital Health Care and Rehabilitation, Swiss Paraplegic Research (SPF), Switzerland 
\AND
\Name{Diego Paez-Granados} 
\Email{diego.paez@hest.ethz.ch} \\ 
\addr SCAI Lab, D-HEST, ETH Zurich, Switzerland\\ 
\addr Digital Health Care and Rehabilitation, Swiss Paraplegic Research (SPF), Switzerland}

\begin{document}

\maketitle

\begin{abstract}
We introduce KarmaTS, an interactive framework for constructing lag-indexed, executable spatiotemporal causal graphical models for multivariate time series (MTS) simulation. 
Motivated by the challenge of access-restricted physiological data, KarmaTS generates synthetic MTS with known causal dynamics and augments real-world datasets with expert knowledge. 
The system constructs a discrete-time structural causal process (DSCP) by combining expert knowledge and algorithmic proposals in a mixed-initiative, human-in-the-loop workflow. The resulting DSCP supports simulation and causal interventions, including those under user-specified distribution shifts. 
KarmaTS handles mixed variable types, contemporaneous and lagged edges, and modular edge functionals ranging from parameterization templates to neural network models. Together, these features enable flexible validation and benchmarking of causal discovery algorithms through expert-informed simulation.
\end{abstract}


\paragraph*{Data and Code Availability}
The code for {KarmaTS} and the interface mentioned in this paper are available at the GitHub repositories
\href{https://github.com/SCAI-Lab/KarmaTS}{KarmaTS} and
\href{https://github.com/SCAI-Lab/KarmaTS-HCI}{KarmaTS-HCI}, respectively.
%This initial paragraph is \textbf{mandatory}. Briefly state what data you
%use (including citations if appropriate) and whether and where the data are
%available to other researchers.
% If you are not sharing code, you must explicitly state that you are not
% making your code available. If you are making your code available, then
% at the time of submission for review, please include your code as
% supplemental material or as a code repository link; in either case, your
% code must be anonymized. If your paper is accepted, then you should
% de-anonymize your code for the camera-ready version of the paper. \emph{If
% you do not include this data and code availability statement for your
% paper, or you provide code that is not anonymized at the time of
% submission, then your paper will be desk-rejected.} Your experiments later
% could refer to this initial data and code availability statement if it is
% helpful (e.g., to avoid restating what data you use).

\paragraph*{Institutional Review Board (IRB)}
This research does not involve human subjects and therefore did not require Institutional Review Board (IRB) approval.

\section{Introduction}

\subsection{Motivation}
Multivariate time series (MTS) arise naturally in domains as diverse as healthcare~\citep{che2018recurrent}, earth science~\citep{angryk2020multivariate}, transportation~\citep{ghosh2009multivariate}, and finance~\citep{tsay2013multivariate}. A hallmark of MTS is the richly interwoven network of temporal and contemporaneous interactions among variables, which causal directed graphs (DAGs), where an edge 
between two variables 
% $X^i \to X^j$ 
%
signifies a direct causal effect can be represented in an interpretable way. 
For instance, causal models built on electronic health records have improved early sepsis prediction~\citep{valik.etal_2023} and characterized recovery trajectories after spinal cord injury~\citep{ehrmann.etal_2020}, while in the geosciences, they have quantified nonlinear dependencies in large observational series~\citep{rungeInferringCausationTime2019}. In engineered systems (e.g., monitoring pin voltages on a circuit board), physical connectivity directly defines the causal graph. 

By contrast, most real-world applications must infer structure from data, often supplementing partial expert knowledge with \emph{causal discovery} algorithms. These range from constraint-based methods~\citep{pc} (PC, FCI) to score-based searches (GES~\citep{chickering2002optimal}) and continuous-optimization approaches (NOTEARS~\citep{notears}). 
%Camera-ready : place a red carpet for expert knowledge
However, the observed statistical distribution is often a noisy or confounded projection of the underlying physiological mechanism—shaped by latent variables, measurement error, and heterogeneous subpopulations—so purely data-driven discovery may miss dependencies that are weak in the data but structurally inevitable from anatomical or mechanistic constraints that domain experts consider unequivocal~\citep{pearl2009causality,rungeInferringCausationTime2019}.

\subsection{Contribution}
We introduce {KarmaTS}, a universal simulation framework that unifies expert knowledge elicitation and algorithmic causal discovery for MTS. Our key contributions are:
\begin{itemize}[noitemsep,topsep=0pt,leftmargin=*]
  \item \textbf{Interactive, lag-indexed graph editor} A user-friendly interface for defining, visualizing, and refining DSCPs (see Definition~\ref{def:dscp}).
  \item \textbf{Executable DSCP runtime} Simulation of MTS from DSCPs, enabling systematic benchmarking and robustness testing of discovery algorithms. 
  \item \textbf{Mixed-initiative human–machine loop} An iterative workflow in which experts and causal discovery algorithms inform one another, leading to progressively more accurate graphical models and higher-fidelity synthetic MTS data.
\end{itemize}

\subsection{Paper Roadmap}
% We demonstrate the contributions of {KarmaTS} in three parts: (i) a feature overview, (ii) a mock privacy-preserving application on fMRI data, and (iii) a suite of synthetic MTS benchmarks for evaluating causal discovery algorithms. By systematically varying factors such as network topology, noise level, and temporal dependence, these benchmarks reveal each method’s strengths and weaknesses and provide actionable guidance for selecting suitable algorithms on comparable real data.

The remainder of this paper is organized as follows. Section~\ref{sec:related work} reviews related work on MTS simulation and causal discovery. Sections~\ref{sec:overview} and~\ref{apdx:interface} formally introduce {KarmaTS} and the accompanying user interface. Section~\ref{sec:example fmri} demonstrates a human–machine loop application of {KarmaTS} on real-world fMRI data. Sections~\ref{sec:benchmark} and~\ref{sec:results} report and analyze empirical evaluations on benchmarks with systematically varied topology and temporal dependence. Finally, Section~\ref{sec:conclusion} discusses limitations and future directions.

\section{Related Work}\label{sec:related work}
Researchers often resort to simulating MTS data to evaluate their algorithms, since high-quality, publicly available datasets remain scarce. While such simulations—built on user-specified dynamics—grant fine-grained control over experimental conditions, they also impose a heavy burden: each new study requires bespoke construction of graphical models and functional mappings. We argue that this redundancy can be greatly reduced by assigning roles more strategically: domain experts should define and validate the causal relationships underlying the data, while algorithm developers concentrate on advancing and benchmarking their methods. The core aim of this work is to bridge these complementary strengths, uniting expert-driven causal specification with streamlined, reusable simulation tools to accelerate and enhance causal discovery research. 

\subsection{Synthetic MTS Generation}

The \textbf{CausalTime} pipeline~\citep{cheng2024causaltime} learns a causal graph from real-world time series using a nonlinear autoregressive model to generate synthetic data. While it uses discriminative scores for fidelity, these aggregate metrics can overlook local patterns and nuances critical in expert-driven fields like healthcare.

\subsection{Interactive Causal Modeling Interfaces}

Several tools support visual or interactive work with causal graphs. 
\textbf{DAGitty}~\citep{dagitty} provides a browser-based editor for static DAGs with adjustment-set analysis and bias diagnostics, and 
\textbf{PyRCA}~\citep{pyrca} offers an interactive environment for root-cause analysis in AIOps, combining causal and dependency graphs with diagnostic workflows.

The classic \textbf{Tetrad} system~\citep{tetrad} combines a GUI for editing and visualizing graphs with a large collection of causal discovery algorithms and basic simulation tools, including time-indexed variables. However, it is organized around generic DAG/SEM and Bayesian network workflows, and time dependence is typically handled by duplicating variables across time slices rather than via a time-series–native notion of lagged and contemporaneous edges with associated functionals.

\subsection{Benchmarking Platforms}

\textbf{CAUSEME}~\citep{rungeInferringCausationTime2019} is a web-based benchmarking platform with curated ground-truth datasets from diverse domains, featuring challenges like non-stationarity and high dimensionality. Its interface allows experts to upload data, select methods, and compare performance metrics (e.g., precision, recall), standardizing offline evaluation to rigorously assess algorithmic robustness.

\subsection{Complementary Roles of {KarmaTS} and CAUSEME}

Whereas CAUSEME focuses on \emph{offline} benchmarking against static and time‐varying ground truths, {KarmaTS} offers an \emph{interactive} simulation environment for \emph{real‐time} definition, modification, and refinement of spatio‐temporal causal graphical models.~Together, these tools form a comprehensive ecosystem. CAUSEME provides community-driven standardized benchmarks for validating causal discovery methods, while {KarmaTS} empowers experts and developers to iteratively design and fine‐tune synthetic MTS datasets tailored to specific research questions. This synergy accelerates both methodological innovation and reliable evaluation in multivariate time series causal discovery. Table~\ref{tab:model_comparison} provides a summary of the comparison.

\begin{table}[h]
\centering
\resizebox{\linewidth}{!}
{
    \begin{tabular}{lccc}
                               & {KarmaTS} & CausalTime & CAUSEME \\ \hline
        \vspace{1px}
        Expert Engagement    &  \cmark    &    \xmark $^1$   &   \xmark \\ 
        \vspace{1px}
        Functional Definition  &    \cmark     &  \xmark  & \xmark  \\ 
        \vspace{1px}
        Realistic              &  \xmark{$^2$}     &   \cmark  &   \cmark   \\ 
        \vspace{1px}
        Real-world Integration & \cmark{$^2$} & \xmark &  \xmark \\ \hline
        &&
    \end{tabular}
}

\caption{Feature comparison between {KarmaTS}, CausalTime, and CAUSEME. }
\label{tab:model_comparison}
\end{table}
Notes about Table~\ref{tab:model_comparison}: 1. While CausalTime can incorporate expert knowledge as a prior graph, this input becomes diluted through the pipeline. 2. Via the human-in-a-loop iteration, {KarmaTS} allows continuous refinement through real-world data fusion to increase the realism. \label{fn:tab2}

\subsection{Causal Discovery in Multivariate Time Series}

Causal discovery in multivariate time series (MTS) addresses temporal and contemporaneous effects, often amid challenges like high dimensionality and irregular sampling.

Constraint-based methods like \textbf{PCMCI}~\citep{pcmci} and its variants (\textbf{PCMCI+}~\citep{pcmci+}, \textbf{LPCMCI}~\citep{LPCMCI}) use conditional independence (CI) tests to filter variables and manage confounders. Structural approaches such as VarLiNGAM~\citep{varlingam} combine vector autoregression with LiNGAM assumptions to find effects in non-Gaussian data.

More recent scalable methods include \textbf{DYNOTEARS}~\citep{dynotears}, which extends the continuous-optimization framework of \textbf{NOTEARS}~\citep{notears} to time series. Deep learning approaches have also emerged: \textbf{NGM}~\citep{ngm2021} uses Neural ODEs to learn causal graphs from irregularly sampled data, while \textbf{TCDF}~\citep{nauta2019causal} applies attention-based networks to find time delays and detect confounders. Graph-neural methods like \textbf{CUTS+}~\citep{cheng2024cuts+} employ GNNs to scale discovery for high-dimensional series.

These diverse approaches offer complementary strengths. For a comprehensive survey of the field, including datasets and evaluation metrics, see Gong et al.~\citep{gong2024causal}.

\section{KarmaTS: An Overview}\label{sec:overview}

{KarmaTS} is a Python library for MTS generation with expert knowledge integration. In this section, we will first introduce the mathematical model along with the assumptions. Following that, we will explain the implementation details.

\subsection{The Mathematical Model}
We used a \emph{discrete-time structural causal process} ~\citep{pcmci+} (DSCP), defined in ~\ref{def:dscp}. All notations are described in Appendix \ref{sec:notation}
\vspace{-5pt}
\begin{definition}\label{def:dscp}
A \emph{discrete-time structural causal process} is a multivariate dynamical system determined by two components: 
\vspace{-5pt}
\begin{enumerate}[
    leftmargin=0.5cm,
    itemsep=0pt,
    parsep=0pt,
    topsep=2pt,
    partopsep=0pt
]
    \item A vector process of 
    $N$ variables $$\mathbf{X}_t = (X_t^1, \ldots, X_t^N)$$
    \item A set of mappings 
    $$X_t^j = f^j\left ( \operatorname{Pa}(X_t^j), \eta_t^j \right ), \quad j = 1, \cdots, N$$ 
\end{enumerate}
where $\operatorname{Pa}(X_t^j) = \left\{X\in \mathbf{X} \mid  \mathbf{X} \in \{\mathbf{X}_t,  \mathbf{X}_{t-1}, \ldots\}\right\} \setminus {X_t^j}$, and $\eta_t^j$ is a random variable from an uncertainty process $\eta^j.$
\end{definition}

\vspace{-5pt}
\begin{remark}
\vspace{-10pt}
\begin{itemize}[
    leftmargin=0.5cm,
    itemsep=0.cm
]
    \item In other parts of this paper, we will refer to the set $\bm{f} = \{f^j \mid j \in [N]\}$  as the functional form of the process.
    \item When not specified otherwise, "parents" indicates the union of the upstream nodes from both the contemporaneous and lagged edges.
\end{itemize}
\end{remark}

\vspace{-20pt}
\begin{definition}
\vspace{-10pt}
       We call an edge $({X}_{t-\tau}^i, {X}_t^j)$ to be 
    \begin{itemize}[leftmargin=0.5cm, itemsep=0pt]
        \item \emph{contemporaneous} if $\tau=0$
        \item \emph{lagged} if $\tau>0$
    \end{itemize}
\end{definition}
\vspace{-5pt}
The DSCP defined is node-oriented, given the time and index of a variable, as depicted in Figure~\ref{fig:DSCP}. Rolling through time, the generation dynamics illustrate a spatial-temporal process whose realization is an MTS. The spatial perspective is due to the mappings across different variables despite their progress in time, and the temporal perspective is due to the existence of lagged edges. Taking clinical monitoring as an example, $\mathbf{X}_t = (X_t^1, \ldots, X_t^N)$ could record the collection of values of human temperature, ECG (electrocardiogram), blood pressure, and so on, at the time step $t$.

\definecolor{myDarkRed}{HTML}{A00000}
\definecolor{myDarkBlue}{HTML}{005588}
\begin{figure}[ht]
  \centering
  
          \tikzset{
        	->,  % makes the edges directed
        	>=stealth, % makes the arrow heads bold
        	shorten >=3pt, shorten <=3pt, % shorten the arrow
                observed/.style={fill=myDarkBlue, text=white, draw=myDarkBlue, thin},
                partially_observed/.style={fill=myDarkRed, text=white, draw=myDarkRed, thin},
                unobserved/.style={fill=gray!60, opacity=0.8, text=black!70, draw=black!50, thin},
                invisible/.style={fill=gray!20, draw=none},
        	node distance=1.2cm, % specifies the minimum distance between two nodes. Change if n
        	every state/.style={fill=blue!20, minimum size=1.2cm, font=\large\ttfamily}, % sets the properties for each ’state’ n
        	initial text=$ $, % sets the text that appears on the start arrow
                blue_edge/.style={myDarkRed, very thick},
                gray_edge/.style={gray!60, thick},
                dash_edge/.style={gray!30, dashed, thick}
         }
     \scalebox{0.6}{
  	\begin{tikzpicture}
            
            \node[state, partially_observed] (xj_t_2) {$X^j_{t-2}$};
            \node[state, unobserved, above=of xj_t_2] (xj_minus1_t_2) {$X^{j-1}_{t-2}$};
            \node[state, unobserved, below=of xj_t_2] (xj_plus1_t_2) {$X^{j+1}_{t-2}$};
            
            \node[state, invisible, left=of xj_t_2] (xj_t_3) {};
            \node[state, invisible, left=of xj_minus1_t_2] (xj_minus1_t_3) {};
            \node[state, invisible, left=of xj_plus1_t_2] (xj_plus1_t_3) {};
            
            \node[state, partially_observed, right=of xj_minus1_t_2] (xj_minus1_t_1) {$X^{j-1}_{t-1}$};
            \node[state, partially_observed, right=of xj_plus1_t_2] (xj_plus1_t_1) {$X^{j+1}_{t-1}$};

            \node[state, unobserved, right=of xj_minus1_t_1] (xj_minus1_t) {$X^{j-1}_t$};
            \node[state, partially_observed, right=of xj_plus1_t_1] (xj_plus1_t) {$X^{j+1}_t$};
            \node[state, observed, below=of xj_minus1_t] (xj_t) {$X^j_t$};

            \node[state, invisible, right=of xj_t] (xj_t_p1) {};
            \node[state, invisible, right=of xj_minus1_t] (xj_minus1_t_p1) {};
            \node[state, invisible, right=of xj_plus1_t] (xj_plus1_t_p1) {};

            \node[state, unobserved, right=of xj_t_2] (xj_t_1) {$X^j_{t-1}$};

            % Gray edges (other relationships)
            \draw[gray_edge] (xj_minus1_t_2) edge[left] node[above] {} (xj_t_1);
            \draw[gray_edge] (xj_plus1_t_2) edge[left] node[above] {} (xj_t_1);
            \draw[gray_edge] (xj_plus1_t_2) edge[left] node[above] {} (xj_t_2);
            \draw[gray_edge] (xj_plus1_t_1) edge[left] node[above] {} (xj_t_1);

            \draw[dash_edge] (xj_minus1_t_3) edge[left] node[above] {} (xj_t_2);
            \draw[dash_edge] (xj_plus1_t_3) edge[left] node[above] {} (xj_t_2);
            \draw[dash_edge] (xj_plus1_t_3) edge[left] node[above] {} (xj_t_3);
            \draw[dash_edge] (xj_t_3) edge[bend left] node[above] {} (xj_t_1);

            \draw[dash_edge] (xj_minus1_t) edge[left] node[above] {} (xj_t_p1);
            \draw[dash_edge] (xj_plus1_t) edge[left] node[above] {} (xj_t_p1);
            \draw[dash_edge] (xj_plus1_t_p1) edge[left] node[above] {} (xj_t_p1);
            \draw[dash_edge] (xj_t_1) edge[bend left] node[above] {} (xj_t_p1);

            \draw[blue_edge] (xj_t_2) edge[bend left] node[above] {} (xj_t);
            \draw[blue_edge] (xj_plus1_t_1) edge[left] node[above] {} (xj_t);
            \draw[blue_edge] (xj_plus1_t) edge[left] node[above] {} (xj_t);
            \draw[blue_edge] (xj_minus1_t_1) edge[left] node[above] {} (xj_t);

  		\node[xshift=-2.5cm] at (xj_t_2) {\texttt{...history}};
            \node[xshift=2.5cm] at (xj_t) {\texttt{future...}};
		\node[yshift=-1.3cm] at (xj_plus1_t_1) {\Large${\color{myDarkBlue}X^j_t} = f^j\left(\,\{\:{\color{myDarkRed}X^j_{t-2}, X^{j-1}_{t-1}, X^{j+1}_{t-1}, X^{j+1}_t}\:\},\: \eta^j\,\right)$};
  
	\end{tikzpicture}
    }
\caption{
An illustration of Definition~\ref{def:dscp}.
}
  \label{fig:DSCP}
\end{figure}
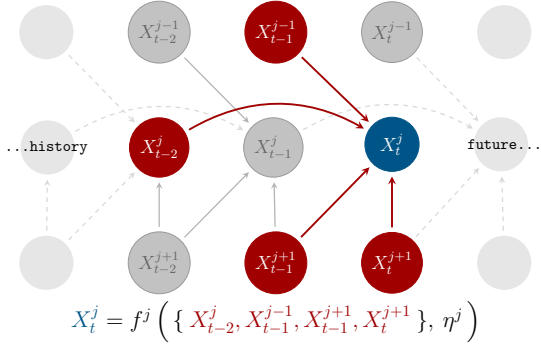

Figure~\ref{fig:DSCP} exemplifies the computation of \(X_t^j\) (blue). Its parents are colored red. The explicit mapping from a set of parent nodes to their common child node must be defined for the simulation. Generally, a node can have multiple parents; therefore, the effects of those parents can be considered together or separately (see Section~\ref {sec:assumption}). Ideally, users should specify this N‐to‐1 mapping function \(f^j\) for each node. Although this is sometimes difficult to achieve, we use a default mapping for each edge if none is specified. In addition, users may define mappings corresponding to non‐intersecting subsets of a node’s parents; in that case, the effects from different parent groups are summed.

% The function \( \bm{f} \) can be complicated depending on the domain of research. 
\paragraph{A Note on the Functionals} The nature of \( \bm{f} \) varies widely across different research domains, taking forms from linear to non-linear, and from simple to intricate. 
This diversity makes it challenging to emphasize any specific paradigm of \( \bm{f} \) as universally representative, which, thankfully, is mitigated by advances in neural networks~\citep{lu2021learning}. DSCP, albeit complicated by the functionals, is mathematically powerful enough for modeling real-world processes~\citep{DBLP:journals/corr/abs-1803-08784}.
In Figure~\ref{fig:fMRI result},~\ref{fig:interfacesim} and \ref{fig:simdata}, we exemplify with synthetic MTS generated by the DSCP.

\subsection{Assumptions}\label{sec:assumption}
\paragraph{Uncertainty Processes} For simplification, we assume the functional $f^j$ is additive with respect to $\operatorname{Pa}(X_t^j)$ and the uncertainty process $\eta_t^j$, which means $X_t^j = f^j\left ( \operatorname{Pa}(X_t^j), \eta_t^j \right ) = f^j\left ( \operatorname{Pa}(X_t^j)\right ) + f^j\left ( \eta_t^j \right )$. Noting that $f^j\left ( \eta_t^j \right )$  is a new uncertainty process, we can further define $g^j_t = f^j\left ( \eta_t^j \right )$. As a result, we land on the model: $$X_t^j = f^j\left ( \operatorname{Pa}(X_t^j)\right ) + g^j_t, \quad j = 1, \cdots, N$$

\paragraph{Additive Decomposition of Edge Functionals}\label{assume_b} The edge function is a mapping from multiple nodes to their common child, i.e., an N-to-1 mapping.  It is not always realistic to specify the joint causal effect in a single attempt. Usually, it is more natural to progressively input 1-to-1 mappings,  or k-to-1 mappings, where k is a considerably smaller number than N. Therefore, we assume $f^j$ is additive with respect to the local causal effects of subsets of parents. To formally model this, we first partition $\operatorname{Pa}(X_t^j)$:$$\operatorname{Pa}(X_t^j) = \bigcup_{k=1}^m \operatorname{Pa}_k
(X_t^j)$$ where $m \leq N$ and $ N=|\operatorname{Pa}(X_t^j)|$. In practice, the partition depends on real-world human interaction. Then,  we define the local causal effects as:
$$f^j_k: \operatorname{Pa}_k(X_t^j) \mapsto {X_t^j}$$
\noindent Finally,  the additive assumption can be written as: 
$$f^j (\cdot) = \sum_{k=1}^mf^j_k(\cdot)$$It is useful to assume $ f^j_k$ to be non-linear for all $k$ because otherwise we can decompose $f^j$ further with a partition of smaller granularity. Additionally, one may notice that when $m=1$, it is equivalent to possessing the N-to-1 joint functionals. 

\subsection{Anatomy of the Framework}
\label{sec:anatony of the framework}
In this section, we will introduce how we programmatically construct the process in the definition ~\ref{def:dscp}. 

\subsubsection*{a. Graphical Model}

Graphical models encode the structure of the DSCP: edges represent the causal relationships among variables. These edges can be (i) derived from empirical expert experience and input via {KarmaTS}’s interface, (ii) learned from data using causal discovery algorithms, or (iii) based on research-backed domain knowledge.

The implementation of the graphical model is built upon NetworkX ~\citep{networkx}, which is a major Python package in network analysis.

\vspace{1px}

\subsubsection*{b. Functional Mappings}
Once the graphical model is defined, a naive mapping is assigned to every edge. The naive mapping can either be an identical mapping, which copies the same value to the successor, or a null mapping, which does nothing to the successor.

% The placeholder mappings tend to create unstable MTS unless human knowledge starts to be incorporated. 
With the additive functional edges assumption mentioned in ~\ref{assume_b}, the human experts will incrementally improve those edges. We provide several options: 

\noindent The following options are available for constructing functional maps, each offering distinct advantages and applications:

\paragraph{i) Parameterized Templates}
These are simple, expert-specified rules for ad-hoc mappings between variable types (e.g., thresholding for a continuous-to-binary edge). Their main advantages are rapid implementation and adaptability to changing requirements.

\paragraph{ii) Neural Network Templates}
This approach uses sophisticated, data-driven models (e.g., ML algorithms, validated equations) to define complex relationships. It offers high reliability and allows for a direct comparison between empirical models and expert knowledge.

\section{User Interface for Expert Knowledge Integration}\label{sec:UI}

\begin{figure}[ht]
\centering
\includegraphics[width=\linewidth, ]{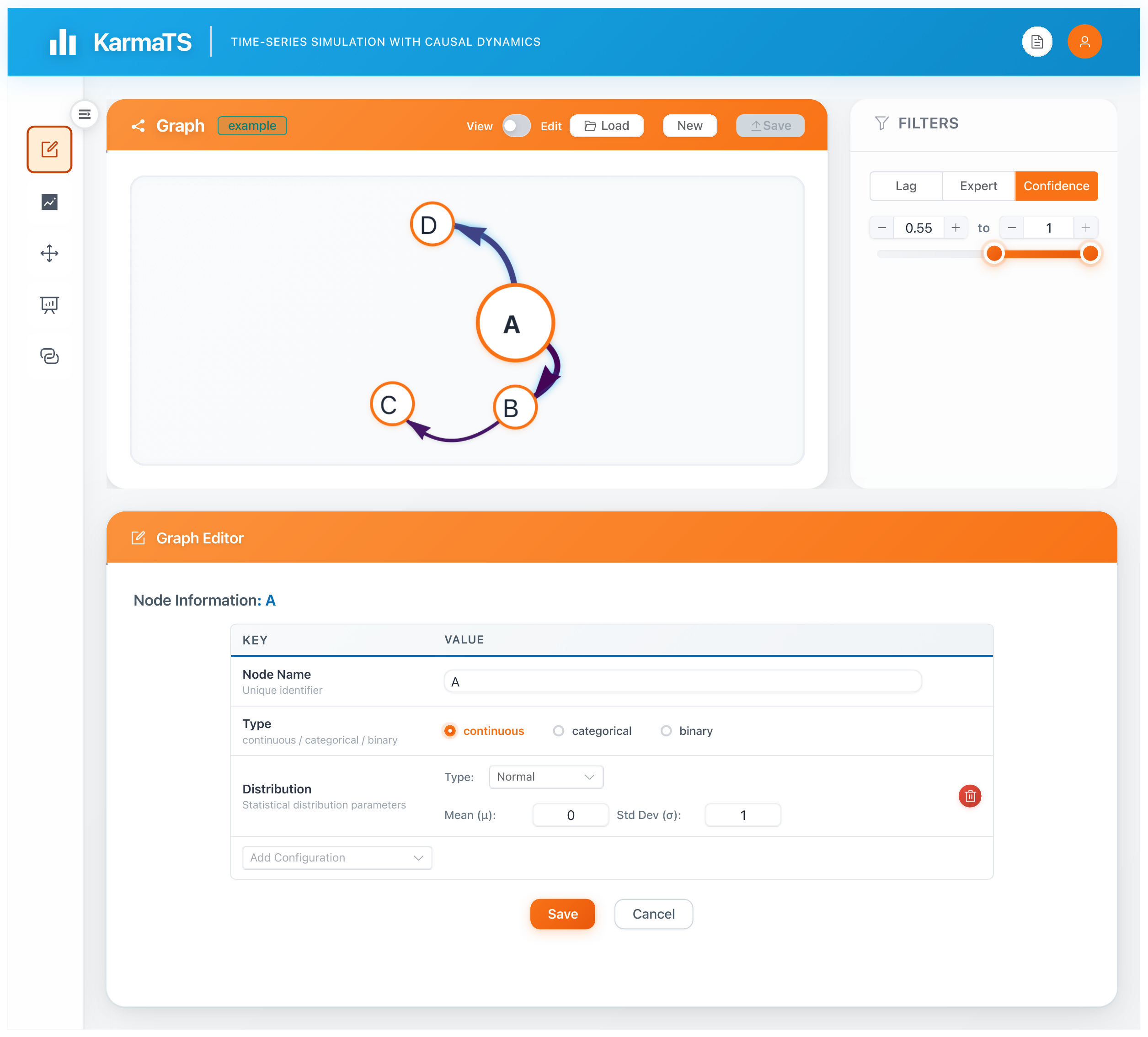}
\caption{
The user interface for expert knowledge input.
% Adding a node is intuitive; clicking the empty area triggers a dialogue for confirmation, which we will show in Figure ~\ref{fig:dialog}. Dragging from one node to a different node will create a temporary edge, which triggers the definition of the edge, including lags, and the aforementioned $f^j_k$ ~\ref{assumptions}. Since edges can be assigned with lags, multiple edges are possible between two variables, as well as self-loop edges for a single variable. Area 3 is an area to preview the generated MTS given a period defined by several time steps, which is 1100 in this example. 
}
\label{fig:UI}
\end{figure}

To facilitate the process of creating, editing, and refining causal graphs, we have developed a user interface, shown in Figure~\ref{fig:UI}. This interface is designed to support researchers and practitioners in the iterative process of causal discovery and model refinement.

The user interface is a comprehensive tool that enables users to construct, visualize, and interact with graphical models for time series data. A more comprehensive introduction of the interface is provided in Section~\ref{apdx:interface}.

\begin{figure*}[ht]
\centering
\subfigure[]{\includegraphics[width=0.3\textwidth]{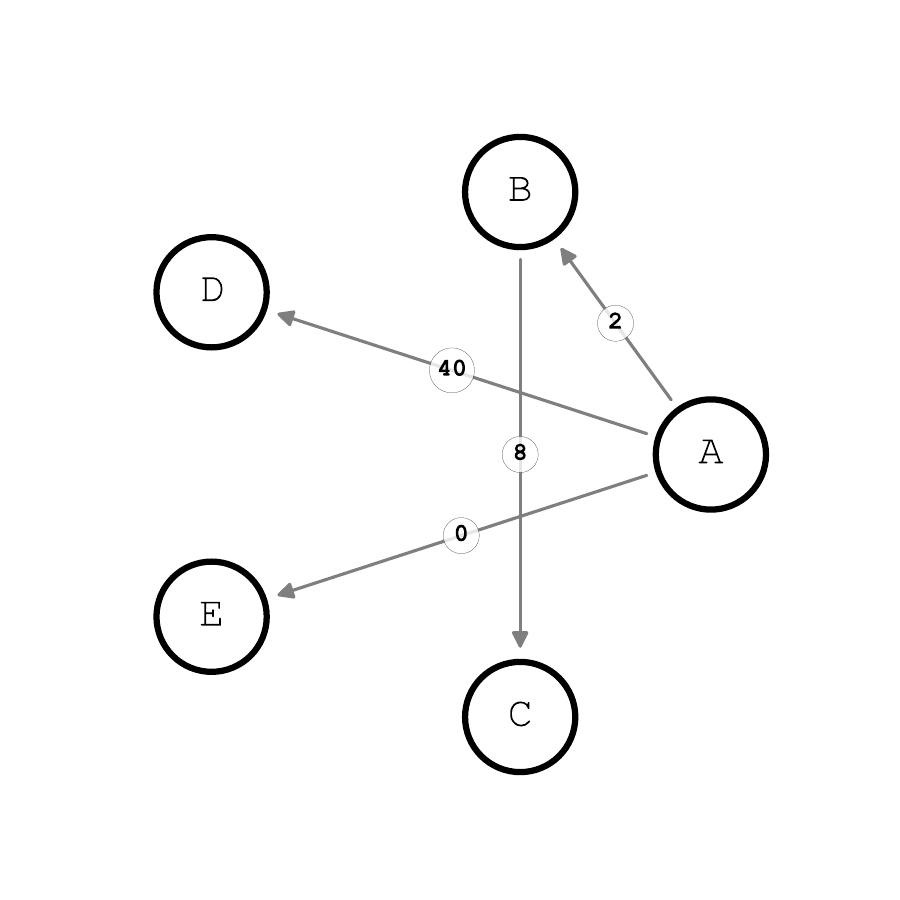}\label{fig:interfacesim:a}}
\subfigure[]{\includegraphics[width=0.65\textwidth]{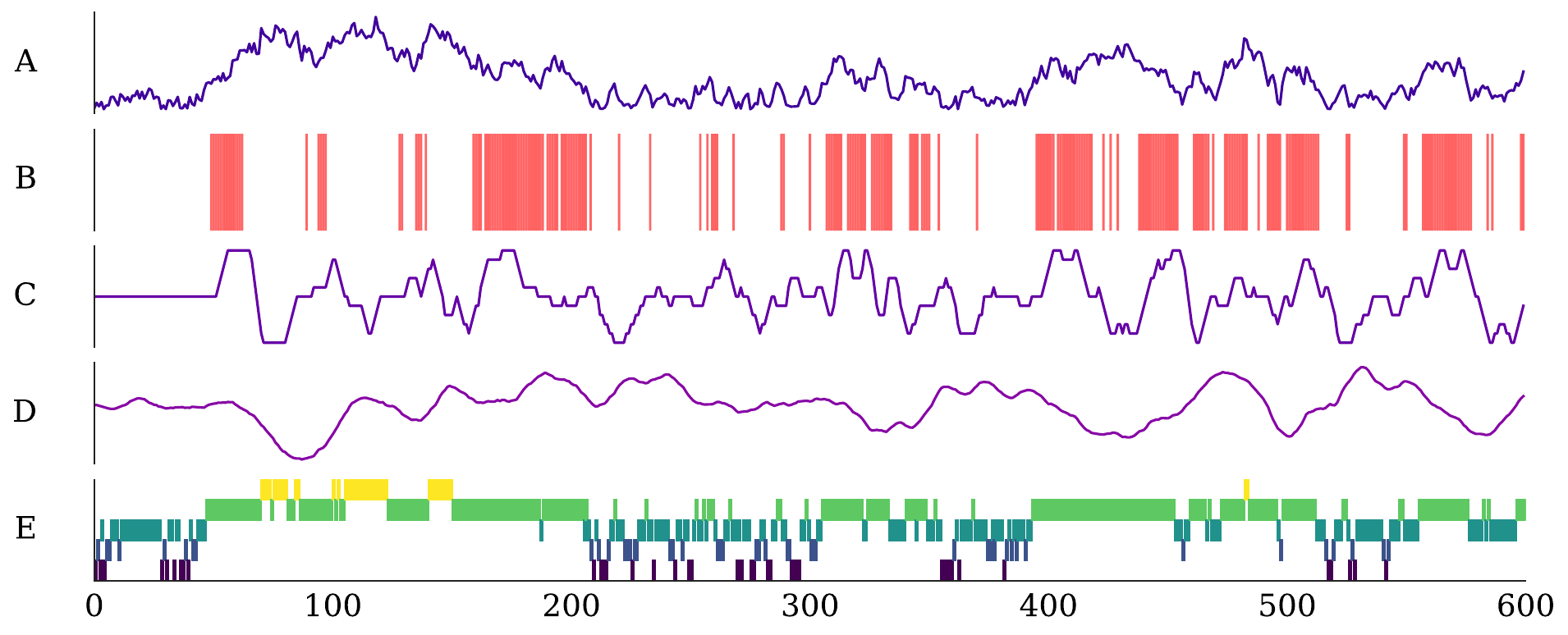}\label{fig:interfacesim:b}}
\caption{
An example of a mixed-type graphical model and the simulated MTS.
}
\label{fig:interfacesim}
\end{figure*}

In Figure~\ref{fig:UI}, the graph view (upper left) is designed to display and edit the graphical model. The image shows a graph created from scratch by an expert, but in practice, it is possible to load graphs from causal discovery algorithms or saved graphs in another workflow. At the bottom is a dialog for the user to specify the details of the graph elements (nodes and edges).

Figure~\ref{fig:interfacesim:a} shows a graphical model created by an expert interacting with the user interface, representing causal relationships between mixed-type variables, including continuous, binary, and categorical variables. The edges indicate the direction and lag of causal influences among the variables. Figure~\ref{fig:interfacesim:b} shows the simulated time series generated based on the graphical model, showing the dynamics of the mixed-type variables. Variable B is binary (red), and variable E is categorical (multi-colored segments), while the remaining variables exhibit continuous values. The time series captures the causal dependencies as defined by the expert-generated model. More interface details can be found in the Appendix~\ref{apdx:interface}.

\subsection{The Assumed Role of Human Experts}
The interface functions as a labeling system that integrates expert insights, blending theoretical knowledge with empirical observations from complex, real-world conditions. For example, clinical experts can use the system to map the nuanced interactions between physiological parameters subject to unpredictable factors like stress or medication.

{KarmaTS} mitigates potential individual bias by valuing the collective knowledge of multiple contributors. The underlying graphical model is refined through a collaborative, iterative process that balances human input with algorithmic adjustments, ensuring a more accurate representation of causal structures.

\subsection{Human-in-the-loop with {KarmaTS}}
\begin{figure}[ht]
\centering
\includegraphics[width=\linewidth, trim={0, 2cm, 0, 0.5cm}]{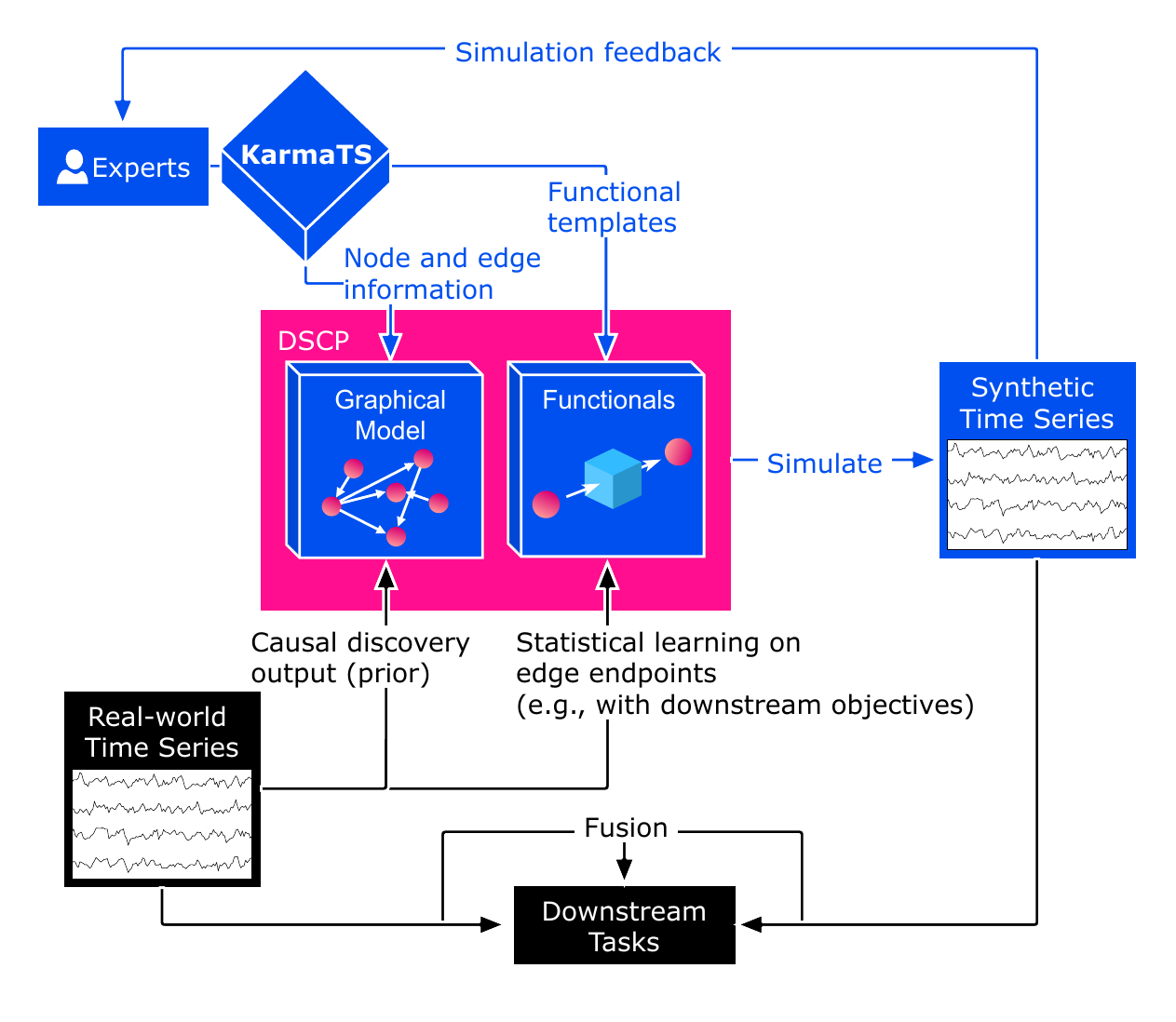}
\caption{
Overview of the human-in-the-loop pipeline with {KarmaTS}. 
}
\label{fig:demo pipeline}
\end{figure}

Figure~\ref{fig:demo pipeline} summarizes the mixed-initiative workflow in {KarmaTS}. %
Experts provide node and edge information, along with initial functional templates, through the interactive editor. In parallel, real-world time series are analyzed by causal discovery algorithms, whose outputs are treated as structural priors, and by statistical learners that fit edge-wise functionals (potentially with losses tied to downstream objectives). Together, the edited graph and learned functionals define a DSCP (Definition~\ref{def:dscp}), which {KarmaTS} then uses to simulate synthetic multivariate time series.%

Synthetic and real-time series can be fused for downstream tasks such as prediction, robustness analysis, or model selection. At the same time, simulated trajectories and diagnostic plots provide feedback to the experts, who may iteratively refine both structure and functionals. 

% Camera Ready: Re-positioning this example. 
\section{Example: Privacy-Conscious fMRI Synthesis}
\label{sec:example fmri}

Figure~\ref{fig:demo pipeline} illustrates how a DSCP can act as a summary of both expert knowledge and real-world time series. In settings where raw time series are access-restricted (e.g., clinical or industrial logs), sharing DSCP parameters or synthetic outputs can offer a more privacy-conscious alternative to distributing the original data.

In this section, we instantiate the workflow in Figure~\ref{fig:demo pipeline} with a small, illustrative, privacy-conscious use case of fMRI data simulation. However, we do not claim any formal privacy guarantees (e.g., differential privacy~\cite{dp_def}), and the actual privacy–utility trade-off will depend on the chosen learning objectives and fusion strategy.
% The goal here is \emph{not} to provide formal privacy guarantees (e.g., differential privacy \cite{dp_def}), but to demonstrate how expert-informed DSCPs can generate synthetic trajectories that preserve network-level structure while decoupling from the exact temporal events observed in the original data. 

\subsection{Dataset and Workflow Overview}
We consider a movie-watching resting-state functional Magnetic Resonance Imaging (fMRI) dataset, using the MSDL (Multi-Subject Dictionary Learning) atlas as implemented in \texttt{Nilearn}~\citep{Richardson2018,Abraham2014}. 

To build the DSCP, we follow the expert-edit route in Figure~\ref{fig:demo pipeline}: the graphical model is constructed by an ``expert'' who relies purely on correlation estimates, and the functionals are modeled by statistical learners (detailed in the next subsection). Concretely, the workflow consists of three components, corresponding to Figure~\ref{fig:demo pipeline}: (i) \emph{real-world time series} given by MSDL-derived fMRI signals loaded from \texttt{Nilearn}, (ii) an \emph{expert knowledge graph} represented by a correlation-thresholded connectivity graph (details in the next subsection), and (iii) \emph{functionals} instantiated as GRU–VAE statistical learners (also detailed below).

\subsection{Instantiating the DSCP for fMRI}
Below, we specify how the two components of the DSCP from Section~\ref{sec:overview} are instantiated in this example.

\paragraph{a) Expert Knowledge (Graphical Model)}
We encode expert knowledge as a correlation-thresholded connectivity graph: nodes are brain regions and an edge is included between regions $i$ and $j$ whenever their functional connectivity $|\mathbf{C}_{ij}|$ exceeds a sparsity threshold $\tau$ following \citet{Zalesky2012}. For each such pair, we add a lag-1 edge $X^i_{t-1} \to X^j_t$, motivated by the low temporal resolution of fMRI and fast neural dynamics, which makes a fixed single-step lag a common simplification \citep{Smith2011, Seth2015}. Self-loops with lag 1 are also included, yielding a DAG
\(
\tilde G
=\bigl(V,E\bigr),\;
E\subseteq\{(X^{i}_{t-1},X^{j}_{t}) : i,j\in[N]\}.
\)
For each node $X^{j}_{t}$, representing a brain region at time $t$, the parent set is
\(
\mathrm{Pa}(X^{j}_{t}) = \{X^{i}_{t-1} : (X^{i}_{t-1},X^{j}_{t})\in E\},
\)
as in Definition~\ref{def:dscp}. This simple, fixed-lag construction does not identify within-lag causal ordering, but provides a concrete anatomy-informed prior for our demonstration.

\paragraph{b) Functionals}
As introduced in Section~\ref{sec:anatony of the framework}, we model the edge-wise functionals with neural networks, using either fidelity-based losses (e.g., reconstruction error) or task-oriented objectives depending on the downstream goal. For each $j\in[N]$ we learn
\[
f^{j} : \mathrm{Pa}(X^{j}_{t})\longrightarrow X^{j}_{t},
\quad
X^{j}_{t} = f^{j}\bigl(\mathrm{Pa}(X^{j}_{t}),\eta^{j}_{t}\bigr),
\]
where $\eta^{j}_{t}$ denotes stochastic innovations. In this example each $f^{j}$ is implemented as a \emph{variational autoencoder} (VAE)~\cite{kingma2013auto} with a \emph{gated recurrent unit} (GRU)~\cite{cho2014learning} encoder for sequence modeling: a GRU encodes the parent trajectories into a latent distribution, from which the VAE samples during training and whose mean is used at inference, and a decoder maps latent codes back to the target outputs. Model structure and training details are given in Appendix~\ref{appendix:fmri model structure} and~\ref{appendix: fmri loss}.

\subsection{Empirical Behavior and Limitations}

\begin{figure}[ht]
    \begin{minipage}[t]{.50\linewidth}
      \vspace{0.5cm}
      \subfigure[]{
        \includegraphics[width=\linewidth]{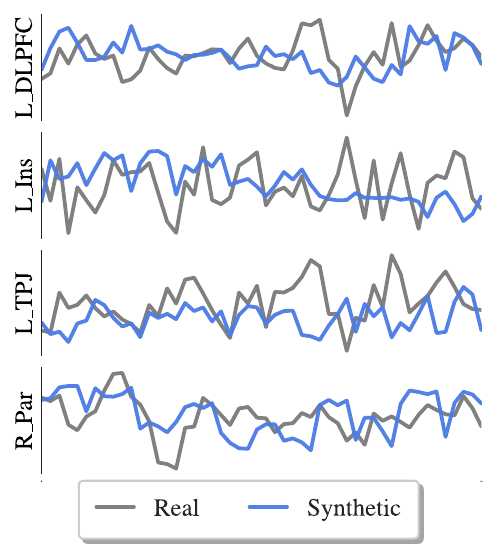}
        \label{fig:subfig1}
      }
    \end{minipage}%
    \hfill
    \begin{minipage}[t]{.49\linewidth}
      \centering
      \subfigure[Real]{
        \includegraphics[width=\linewidth]{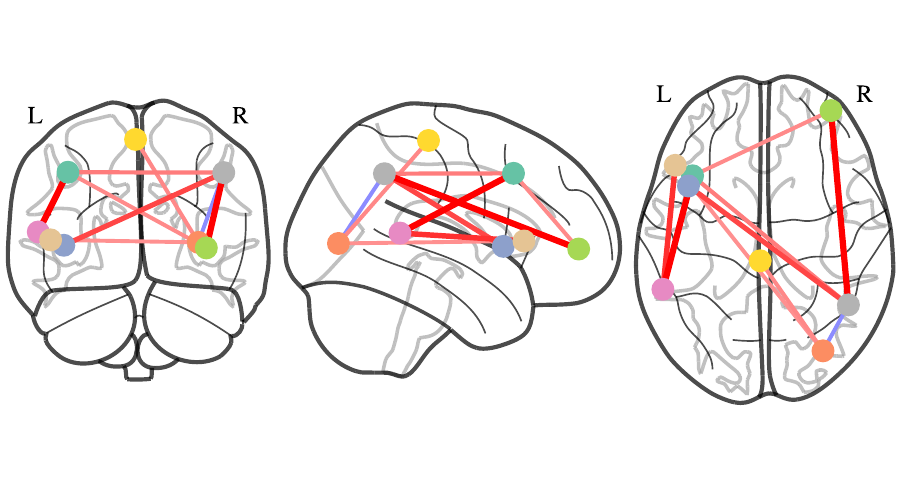}
        \label{fig:subfig2}
      }
      \subfigure[Synthetic]{
        \includegraphics[width=\linewidth]{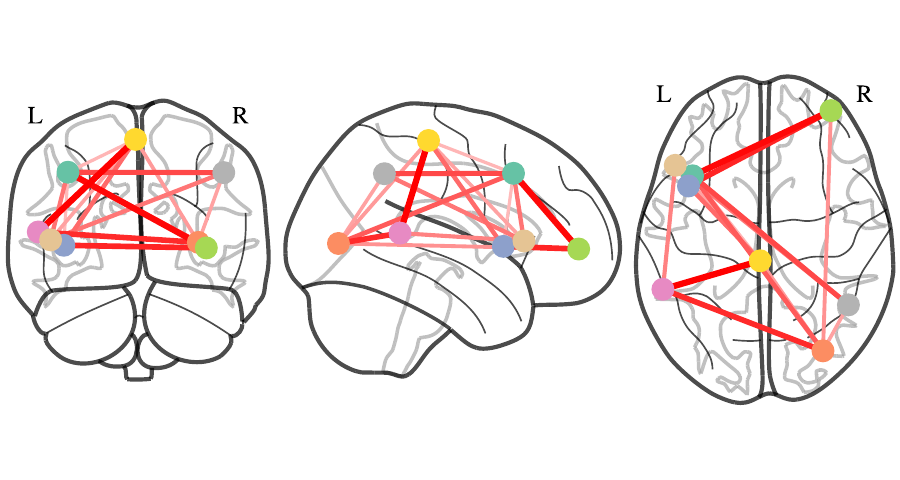}
        \label{fig:subfig3}
      }
    \end{minipage}%
    \caption{
The outcome of privacy-conscious synthetic data generation. We also provide a quantitative comparison in Appendix~\ref{appendix:fmri more results}.}
    \label{fig:fMRI result}
\end{figure}

In Figure~\ref{fig:subfig1}, the synthetic MTS (blue) is initialized with a short segment of real fMRI data (gray) to match the initial state, but subsequently diverges from the real events, thereby reducing the risk of reproducing subject-specific temporal patterns. Figures~\ref{fig:subfig2} and~\ref{fig:subfig3} compare the functional connectivity networks derived from the real and synthetic data, respectively. 

Qualitatively, the synthetic network preserves the global organization of the real map:
(i) strong left–right mirror–pair edges remain dominant over other cross-hemispheric links,
consistent with homotopic interhemispheric functional connectivity~\citep{Zuo2010};
(ii) the large-scale community structure resembles canonical resting-state networks reported by
\citet{Yeo2011,Power2011};
(iii) key long-range hub pathways align with known hub/rich-club architecture~\citep{vandenHeuvel2011};
and (iv) although local intra-hemispheric links differ and some weights are redistributed
(e.g., a somewhat stronger midline hub, slightly weaker ipsilateral columns), bilateral mirror cores
remain salient. This suggests that the synthetic data can maintain key network-level properties of the original dataset while obfuscating fine-grained temporal details. We additionally report a quantitative comparison in Appendix~\ref{appendix:fmri more results}.

% The synthetic data reproduces the overall network structure and major connections observed in the real data, while exhibiting discrepancies in correlation strength (with darker lines indicating stronger connections). 

We emphasize that this example is an illustrative, semi-synthetic use case, not a state-of-the-art privacy mechanism: we do not claim formal privacy guarantees such as differential privacy, and the fidelity–privacy trade-off will depend on the choice of functionals, loss, and fusion strategy.

% \subsection{Benefits}
% The interface is an essential component of our proposed pipeline, providing a seamless and efficient way for researchers to engage in the causal discovery process. By combining algorithmic learning with human expertise, the interface ensures that the resulting causal graphs are both data-driven and expert-informed. The ability to generate synthetic data and iteratively refine models enhances the robustness and applicability of the causal discovery process. This tool reduces the complexity and effort involved in causal graph construction and refinement, enabling researchers to focus on deeper analysis and interpretation of causal relationships in MTS data.

\section{Benchmarking Causal Discovery Algorithms on {KarmaTS}}\label{sec:benchmark}

Standard benchmarks in causal discovery often rely on generic, algorithmically generated graphs that lack domain-specific realism. Involving domain experts is crucial, as their labeled models capture important empirical nuances and can serve as privacy-preserving summaries of sensitive data. This enables the creation of safer, more meaningful benchmarks that reflect real-world dynamics.

\paragraph{Causal Discovery for Time Series Data} Causal discovery algorithms have varied assumptions (e.g., constraint- vs. gradient-based), leading to divergent results with no standard evaluation metric. The best measure of performance, therefore, is an algorithm's ability to reconstruct a known ground truth graph. We evaluate how well different methodologies recover template-based, user-specified ground-truth graphs, which serve as proxies for expert-defined structures. We will evaluate the algorithms mentioned in Section~\ref{sec:related work} using mainstream metrics, i.e., the F1-score and Structural Intervention Distance (SID), on a simulated dataset generated with a specific configuration. Details of the metrics and the simulated dataset are described in Appendix~\ref{apdx:benchmark}.

\section{Results and Discussion}\label{sec:results}

We report empirical findings in three stages.  
First, Section~\ref{sec:overall_perf} summarizes \emph{overall accuracy} of six state-of-the-art (SOTA) causal discovery algorithms across the full suite of {KarmaTS} benchmarks.  
Section~\ref{sec:factor_analysis} disentangles how specific data-generating factors—series length, graph topology, edge density, maximum lag, and latent variable rate—drive performance differences. 

\paragraph{Demonstration Protocol}\label{misc:plot}  
Metrics are plotted as line charts against contextually chosen variables (e.g., time horizon and node count). F1-score is presented with an error band denoting the standard deviation across repeated runs with different random seeds. These error bands are symmetric. 

\subsection{Overall Performance Analysis}
\label{sec:overall_perf}

\begin{figure}[h]
        \centering
        \includegraphics[width=\linewidth]{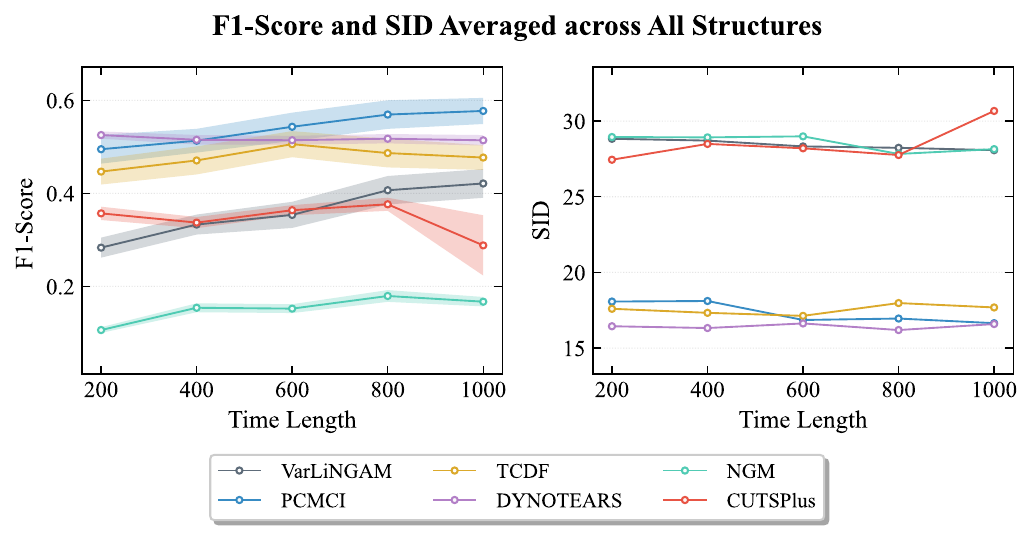}
        \caption{
        Performance of six causal discovery methods as time length increases (200–1000), averaged over all structures (see~\ref{sec:sim_data}). 
        }
        \label{fig:exp-samplesize}
\end{figure}

Figure~\ref{fig:exp-samplesize} reveals a clear distinction in model performance across both metrics. Based on the F1-Score, \textbf{PCMCI} emerges as the top performer at longer time lengths (F1-Score $\sim$0.58), while \textbf{DYNOTEARS} demonstrates the most stability ($\sim$0.52 across all lengths). This is reinforced by the SID analysis, where \textbf{DYNOTEARS} is the clear winner with the lowest and most stable error rate (SID $\sim$16.5). In contrast, \textbf{NGM} is consistently the weakest performer in terms of accuracy (F1-Score $<$ 0.2), and \textbf{CUTS+} shows degraded performance with high error rates, especially with longer time series.

\paragraph{External calibration to prior benchmarks}
This relative ordering is broadly consistent with established time-series causal discovery benchmarks such as CAUSEME and related studies~\citep{rungeInferringCausationTime2019,gong2024causal,pcmci,dynotears,nauta2019causal,ngm2021,cheng2024cuts+}: constraint-based methods like \textbf{PCMCI} and continuous-optimization approaches like \textbf{DYNOTEARS} typically achieve the strongest structural accuracy, while deep models such as \textbf{TCDF} and \textbf{NGM} exhibit more variable performance. Our {KarmaTS} results align with these trends at a coarse level and further clarify how performance shifts with lag, density, and latent-variable rate; Appendix~\ref{apdx:external_calibration} provides a brief mapping to prior benchmark reports.

\subsection{Factor–wise Performance Analysis}
\label{sec:factor_analysis}
Although certain models achieve strong overall performance, some algorithms demonstrate clear advantages in specific datasets. {KarmaTS} enables researchers to evaluate performance under more granular and adaptable configurations, offering deeper and more targeted insights.

\begin{figure}[h]
    \centering
    \includegraphics[width=\linewidth]{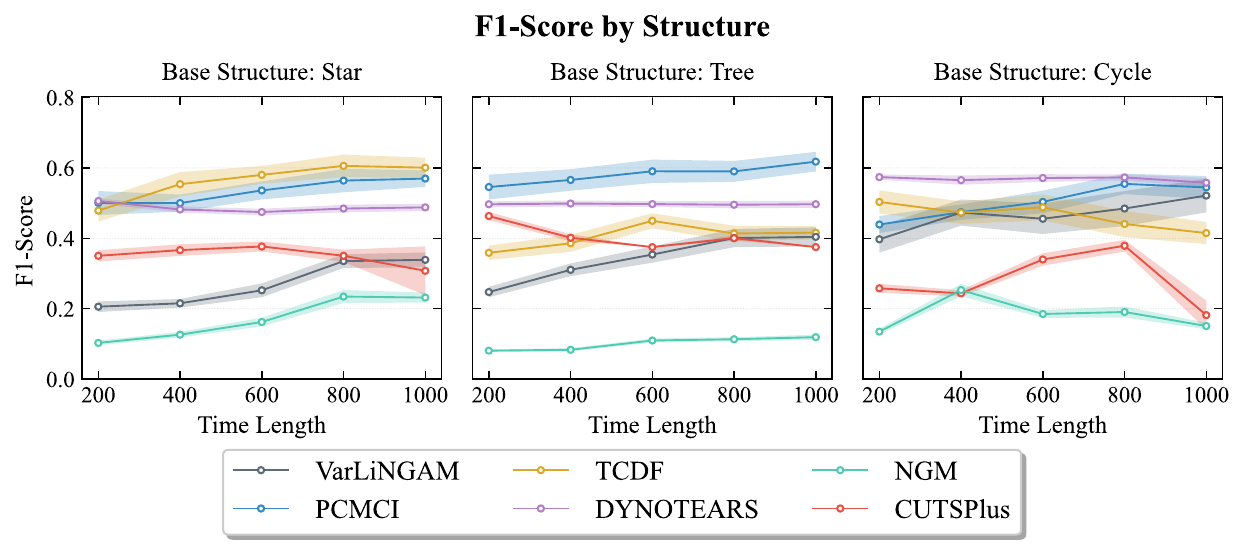}
    \caption{
    Average F1-scores for three prototypical graph structures (Star, Tree, Cycle). Results have been averaged over all edge configurations (see Section~\ref{sec:sim_data}).
    }
    \label{fig:exp-struct}
\end{figure}

As an example, Figure~\ref{fig:exp-struct} demonstrates that no single model is universally superior. Instead, algorithms exhibit clear, context-specific advantages. For the \textbf{Star} structure, \textbf{TCDF} shows a distinct advantage, peaking with the highest F1-Score of approximately 0.62. On \textbf{Tree} structures, \textbf{PCMCI} is the dominant performer, maintaining a stable F1-Score of $\sim$0.6. In the more complex \textbf{Cycle} configuration, both \textbf{PCMCI} and \textbf{DYNOTEARS} prove most robust, achieving stable F1-Scores around 0.55. This targeted evaluation allows researchers to gain finer insights beyond aggregate performance metrics and understand the precise conditions under which a specific algorithm will succeed or fail.
More factor-wise analysis is organized in Appendix~\ref{apdx:result}.

% \subsection{Limitations and Generalizations}

% \paragraph{Limitations}
% Our evaluation uses synthetic data only; validating on real MTS is future work
% (e.g., beyond simulator-based benchmarks such as CauseMe \citep{CauseMe2020}).
% {KarmaTS} provides a configurable sandbox for stress-testing causal
% discovery \citep{pcmci,LiNGAM,dynotears,TCDF}.

% \paragraph{Causal discovery on tabular data}
% The discrete-time structural causal process reduces to a contemporaneous causal
% graph when the lag order is zero (cf. SCMs \citep{pearl2009causality,
% Peters2017Elements} and VAR/SVAR models \citep{Lutkepohl2005VAR}). Treating the
% series as tabular data is possible by destroying temporal order (e.g., random
% permutation), yielding i.i.d.-like samples \citep{Peters2017Elements}.

% \paragraph{Time-series forecasting and imputation}
% Synthetic MTS is widely used to develop and evaluate forecasting and imputation
% methods \citep{
% Hyndman2018Forecasting,
% Makridakis2018M4,
% caoBRITSBidirectionalRecurrent2018,
% GAIN2018,
% GPVAE2020}, but simple random-graph generators with short lags may miss complex spatio‑temporal
% dependencies common in real systems \citep{TimeSeriesSurvey2021}.

\section{Conclusion}\label{sec:conclusion}
{KarmaTS} offers a useful tool for integrating human expertise with algorithmic processes for causal discovery in MTS data. 
By supporting collaboration between domain experts and computational models, the system enhances the representation of causal dynamics. With its flexible interface, customizable uncertainty features, and the ability to simulate datasets with known ground truth, {KarmaTS} provides a valuable resource for research in various fields, including physiology. Additionally, the system contributes to the iterative refinement of causal models and the evaluation of causal discovery algorithms.

\paragraph{Limitations} 
The system lacks comprehensive validation of its effectiveness across user groups and use cases, including medical researchers, data scientists, and domain experts, and the benchmarks used in this paper currently rely on template-based, user-specified proxy graphs rather than fully elicited expert structures. More extensive testing is needed to evaluate how different users interact with and benefit from the system's features, particularly in real-world analytical scenarios.

\paragraph{Future Work}
Future research directions include conducting user studies with domain experts to validate the system's effectiveness and usability, as well as refining the expert-elicitation protocol used to construct and revise DSCP graphs. Additionally, we aim to incorporate more sophisticated statistical methods for uncertainty estimation and develop robust quantification approaches for learned causal relationships. These improvements will strengthen the system's support of reliable causal inference in time series analysis.

\section*{Acknowledgment}
This research project was partially supported by the JST Moonshot R\&D Program, Grant Number JPMJMS2034-18.

\bibliography{references, documentations, extra}

\newpage

\appendix

\section{Notations} \label{sec:notation}
We denote the set of characteristics by $X^j$, where $j \in [N]$ and $N$ is the number of characteristics. Each feature $X^j$ also represents the associated random process $X^j: t \mapsto X_t^j$, mapping time steps to random variables. Following standard conventions, we use $X_t^j$ to denote the random variable at time step $t$, and $x_t^j$ for its observed value (realization).

In our graphical model, the nodes correspond to the random variables $X_t^j$. Since there is no ambiguity, we use $X_t^j$ to refer to both the random variable and the corresponding node in the graph. The set $\operatorname{Pa}(X_t^j)$ denotes the causal parents of $X_t^j$ within the graph. Extending this notation to realizations, $\operatorname{Pa}(x_t^j)$ represents the collection of observed values corresponding to the parents of $X_t^j$.

For edge notation, we use $(X, Y)$ to represent a directed edge from $X$ to $Y$, indicating that $X$ is a direct cause of $Y$. An undirected edge between $X$ and $Y$ is denoted by $\{X, Y\}$.

\section{Supplementary Information for the fMRI Example} \label{appendix:training details}
\newcommand{\Pa}{\operatorname{Pa}}

\subsection{VAE-based Functional Model}\label{appendix:fmri model structure}

For a given target process $X^j$ with parent set $\Pa(X^j)$, we model the edge functional from parent histories to the next value of $X^j$ using a variational autoencoder (VAE)~\citep{kingma2013auto} with a GRU encoder.

Fix a history length $L$ and a prediction horizon $H$. For each time index $t$, we collect the parent history window
\[
\Pa(x_{t-L+1:t}^j)
\;\in\; \mathbb{R}^{L \times d_j},
\]
where $d_j = \lvert \Pa(X^j) \rvert$ is the number of parents, and the corresponding target sequence for the child
\[
y_{1:H}^j
\;=\;
\bigl(x_{t+1}^j,\dots,x_{t+H}^j\bigr).
\]

The encoder is a unidirectional GRU that reads the parent sequence and outputs a hidden state $h^j \in \mathbb{R}^{h}$ summarizing the history:
\[
h^j
\;=\;
\mathrm{GRU}_\phi\bigl(\Pa(x_{t-L+1:t}^j)\bigr).
\]
From $h^j$ we obtain the mean and (log-)variance of a latent Gaussian,
\[
\mu^j \;=\; W_\mu h^j + b_\mu,
\qquad
\log \sigma^{j2} \;=\; W_\sigma h^j + b_\sigma,
\]
and sample a latent code via the reparameterization trick
\[
z^j
\;=\;
\mu^j \;+\; \sigma^j \odot \varepsilon,
\qquad
\varepsilon \sim \mathcal{N}(0, I).
\]
In our implementation, the latent code $z^j$ is directly projected to a scalar prediction of the next value of $X^j$; rolling this one-step predictor forward with teacher forcing on non-target variables yields a predicted sequence
\[
\hat{y}_{1:H}^j
\;=\;
\bigl(\hat{x}_{t+1}^j,\dots,\hat{x_{t+H}^j}\bigr)
\]
that is used in the DSCP update for node $X^j$.

\subsection{VAE Mixed Loss}\label{appendix: fmri loss}

Training uses a composite loss that balances sequence reconstruction, low-order statistics, temporal dependence, cross-variable correlation structure, and a KL regularizer on the latent space. For brevity, we drop the superscript $j$ and write $y_{1:H}$ and $\hat{y}_{1:H}$ for the true and predicted sequences over a batch.

\paragraph{Reconstruction Term}
We define a time-weighted mean-squared error over the prediction horizon:
\[
\mathcal{L}_{\mathrm{rec}}
=
\frac{1}{B}
\sum_{b=1}^B
\sum_{h=1}^H
w_h\,
\bigl(\hat{y}_{b,h} - y_{b,h}\bigr)^2,
\]
where $B$ is the batch size and $w_h$ increases with $h$ to emphasize later predictions.

\paragraph{Marginal Statistics}
To encourage the synthetic trajectories to match the marginal level and variability of the real series, we match the global mean and standard deviation over batch and time:
\[
\mathcal{L}_{\mathrm{stat}}
=
\bigl(\hat{\mu} - \mu\bigr)^2
+
\bigl(\hat{\sigma} - \sigma\bigr)^2,
\]
where
$\hat{\mu},\hat{\sigma}$ are the mean and standard deviation of $\hat{y}_{1:H}$ and
$\mu,\sigma$ are those of $y_{1:H}$.

\paragraph{Lag-1 Autocorrelation}
We also match the lag-1 autocorrelation,
\[
r
=
\mathbb{E}[y_{t} y_{t+1}],
\qquad
\hat{r}
=
\mathbb{E}[\hat{y}_{t} \hat{y}_{t+1}],
\]
via
\[
\mathcal{L}_{\mathrm{ac}}
=
\bigl(\hat{r} - r\bigr)^2,
\]
computed empirically over batch and time.

\paragraph{Cross-variable Correlation Row}
Let $X_t^1,\dots,X_t^N$ denote all processes at time $t$ and consider the empirical Pearson correlations between $X^j$ and all other variables over the horizon, arranged in a vector
\[
c \in \mathbb{R}^N,
\qquad
c_k = \operatorname{corr}\bigl(y_{1:H}, x_{1:H}^k\bigr).
\]
Using the same construction with $\hat{y}_{1:H}$ yields $\hat{c} \in \mathbb{R}^N$. We penalize deviations in this correlation “row” via
\[
\mathcal{L}_{\mathrm{corr}}
=
\lVert \hat{c} - c \rVert_2^2.
\]

\paragraph{KL Regularization}
For each latent code, the encoder defines a Gaussian posterior $q_\phi(z \mid \Pa(x_{t-L+1:t}^j)) = \mathcal{N}(\mu,\operatorname{diag}(\sigma^2))$. We regularize it toward the standard normal prior $p(z) = \mathcal{N}(0,I)$ using the closed-form KL divergence
\[
\mathcal{L}_{\mathrm{KL}}
=
\mathbb{E}\Biggl[
\frac{1}{2}
\sum_{d}
\Bigl(
\mu_d^2 + \sigma_d^2 - 1 - \log \sigma_d^2
\Bigr)
\Biggr],
\]
averaged over batch and time.

\paragraph{Total Objective}
The full VAE mixed loss combines these terms with scalar weights
\begin{align*}
\mathcal{L}_{\mathrm{VAE}}
  &= \mathcal{L}_{\mathrm{rec}}
   + \lambda_{\mathrm{stat}} \mathcal{L}_{\mathrm{stat}}
   + \lambda_{\mathrm{ac}}   \mathcal{L}_{\mathrm{ac}} \nonumber\\
  &\quad
   + \lambda_{\mathrm{corr}} \mathcal{L}_{\mathrm{corr}}
   + \lambda_{\mathrm{KL}}   \mathcal{L}_{\mathrm{KL}}.
\end{align*}
where $\lambda_{\mathrm{stat}},\lambda_{\mathrm{ac}},\lambda_{\mathrm{corr}},\lambda_{\mathrm{KL}}$ control the strength of each regularizer. This objective encourages each learned edge functional to reproduce not only the one-step dynamics of $X^j$, but also its basic marginal behavior, short-range temporal dependence, and correlation pattern with the rest of the MTS, while maintaining a regularized latent representation.

\subsection{Results} \label{appendix:fmri more results}
\begin{figure} \label{fig:fmri corr}
    \centering
    \includegraphics[width=\linewidth]{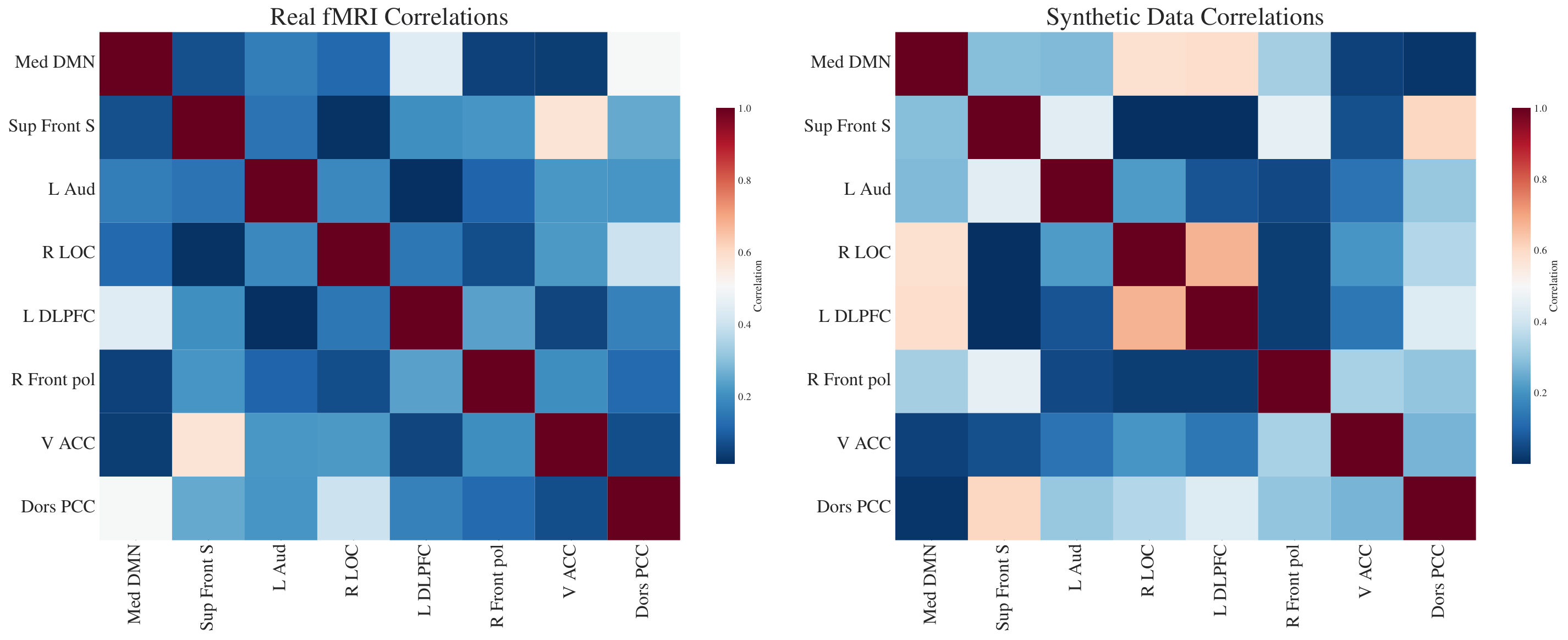}
    \caption{Correlation matrix of the brain connections from real fMRI data (left) and synthetic time series (right).}
    \label{fig:placeholder}
\end{figure}

To quantify pattern similarity between real and synthetic connectivity, we report
a ``matrix correlation'' (\texttt{Matrix\_Corr}) score: the Pearson correlation
between the vectorized off-diagonal entries of the two Pearson correlation
matrices, following common practice in network neuroscience for comparing
connectomes~\citep{Yeo2011,Power2011,vandenHeuvel2011}. In
Figure~\ref{fig:fmri corr}, this score is $0.5097$, indicating moderate alignment
of the edge-strength patterns despite differences in individual weights.

\subsection{Same Expert Graph with Different Functionals}

\paragraph{Setup} We keep the same expert graph and swap the edge functionals: (i) a VAE-based generator and (ii) a Transformer-based generator. We then compare the synthetic vs. real Pearson correlation matrices in Table~\ref{tab:comparison} using the following complementary metrics to measure the similarity between brain connection graphs from synthetic data and real data.

\begin{table}[h]
    \centering
    \begin{tabular}{c|c|c}
        Metrics & VAE & Transformer \\ 
        \hline
        Matrix\_Corr ($\uparrow$) & \textbf{0.5097} & -0.0342 \\
        MAE ($\downarrow$) & 0.3820 & \textbf{0.2993} \\
        RMSE ($\downarrow$) & 0.4927 & \textbf{0.4355} \\ 
        Frobenius Norm ($\downarrow$) & 3.6871 & \textbf{3.2592} \\
        Cosine ($\uparrow$) & \textbf{0.5178} & 0.1015 \\ 
        Spectral $L_2$ ($\downarrow$) & 1.6243 & \textbf{0.6020}
    \end{tabular}
    \caption{Matrix\_Corr and Cosine similarity measure the overall pattern and edge directional alignment. MAE, RMSE, and Frobenius Norm showcase the absolute entry-wise distance, while Spectral $L_2$ tells the global eigen-structure similarity via the $L_2$ distance between eigenvalue spectra.}
    \label{tab:comparison}
\end{table}

\paragraph{Observation} There is a clear trade-off: the VAE better preserves the pattern of correlations (relatively strong/weak edges), while the Transformer is closer in absolute error and eigen-structure. For uses that threshold correlations to recover edges/hubs or compare top-k links, the VAE is preferable (higher Matrix\_Corr/Cosine). For applications driven by global network statistics (e.g., spectral or diffusion properties), the Transformer is preferable (lower MAE/RMSE/Frobenius Norm and Spectral $L_2$). 

\paragraph{Conclusion} 
This is a small, illustrative experiment, not a state-of-the-art benchmarking effort. The comparison in the table is intended to clarify behavior under a fixed modeling choice, not to rank models. 

Within this constrained setup, the table highlights a trade-off: the VAE better preserves pattern alignment, while the Transformer is closer in absolute error and spectral structure. This enriches the example without positioning it as a competitive benchmark.

\section{User-Interface}\label{apdx:interface}

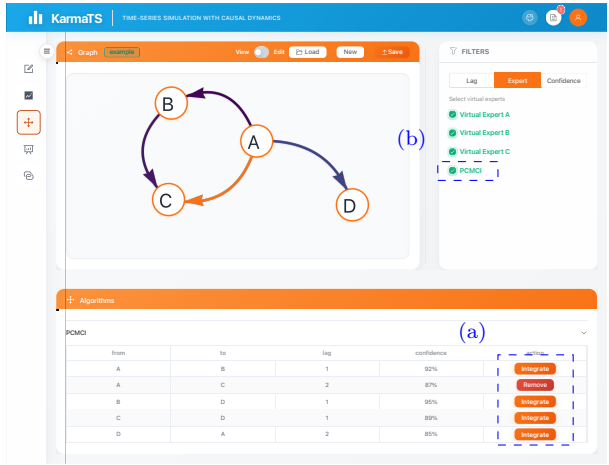
\begin{figure}
    \centering
    \begin{tikzpicture}
        % the image
        \node[inner sep=0] (img) {\includegraphics[width=\linewidth]{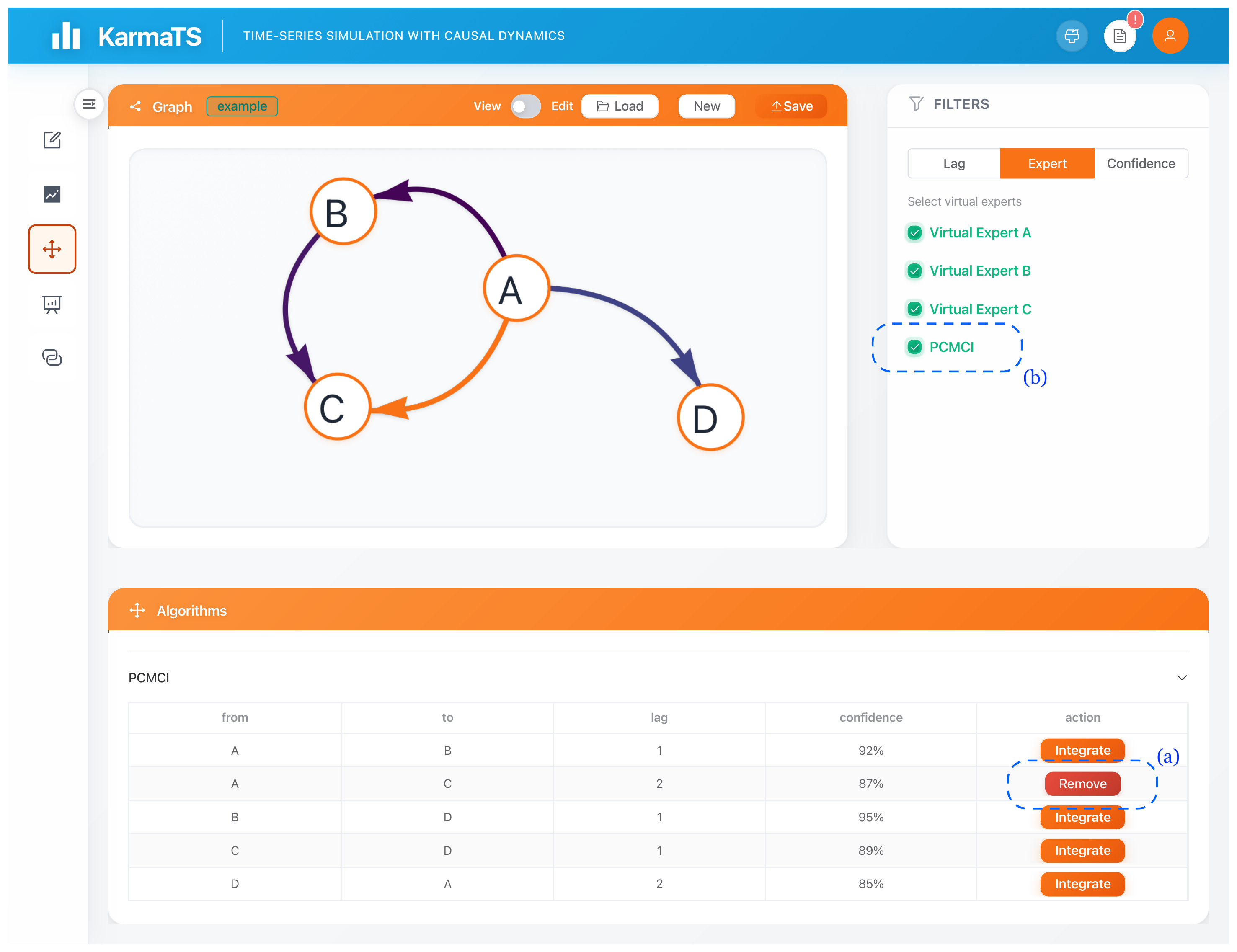}};
        
        % ---- Box (a) : upper-right region ----
        \node[
            draw=blue,
            dashed,
            fill=none,
            anchor=north east,
            xshift=-15mm,
            yshift=-21.5mm,
            inner sep=0pt,
            minimum width=0.8cm,
            minimum height=0.25cm
        ] (boxa) at (img.north east) {};

        % label (a) upper-left of box
        \node[
            anchor=south east,
            text=blue,
            font=\footnotesize,
            xshift=-.5pt,
            yshift=.5pt
        ] at (boxa.north west) {(b)};
        
        % ---- Box (b) : lower-right region ----
        \node[
            draw=blue,
            dashed,
            fill=none,
            anchor=south east,
            xshift=-5mm,
            yshift=3mm,
            inner sep=0pt,
            minimum width=1cm,
            minimum height=1.2cm
        ] (boxb) at (img.south east) {};

        % label (b) upper-left of box
        \node[
            anchor=south east,
            text=blue,
            xshift=-.5pt,
            font=\footnotesize,
            yshift=.5pt
        ] at (boxb.north west) {(a)};
    \end{tikzpicture}
    \caption{Edges suggested by causal discovery algorithms. (a) Experts can decide which ones to integrate. (b) An ``expert'' named after the algorithm will appear for the integrated edge. }
    \label{fig:ui cd edge}
\end{figure}

\subsection{User Interface Features}\label{uif}
\begin{figure}
    \centering
    \includegraphics[width=\linewidth]{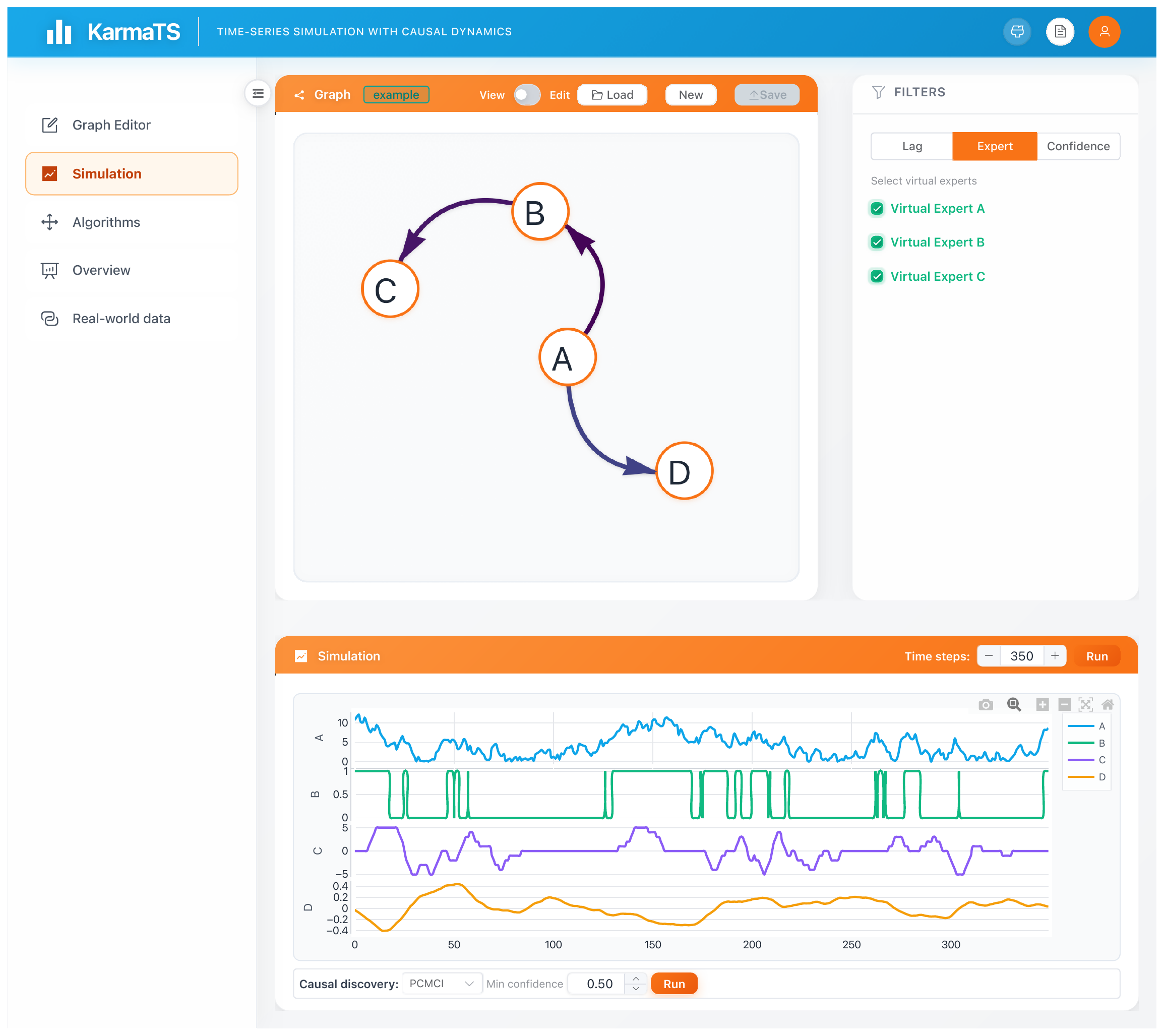}
    \caption{Time series simulation from a defined DSCP over a set of mixed-type variables.}
    \label{fig:ui sim}
\end{figure}

Key functionalities of the interface include:
\paragraph{Graph Creation and Visualization} As shown in the upper left panel in Figure~\ref{fig:ui cd edge}, users can interactively build, edit, and visualize causal graphs alongside their corresponding time series data, labeling nodes and edges with domain-specific meanings.
\paragraph{Algorithm Integration} The interface integrates with causal discovery algorithms to generate initial graphs, which experts can then iteratively refine based on their domain knowledge (see Figure~\ref{fig:ui cd edge}).
\paragraph{Synthetic Data Generation} Users can generate configurable synthetic time series data from the specified causal graphs to test hypotheses and validate causal models (see Figure~\ref{fig:ui sim}).
\paragraph{Collaborative Environment} The system supports real-time, multi-user editing of causal graphs, with version control included to track and manage changes.

% in the preamble:
% \usepackage{tikz}
% \usetikzlibrary{positioning}

\subsection{Example Editing View}
\begin{figure}[h]
\centering
\subfigure[Node editor for categorical variables]{\includegraphics[width=1\linewidth, trim={10cm 1cm 1cm 26.2cm}, clip]{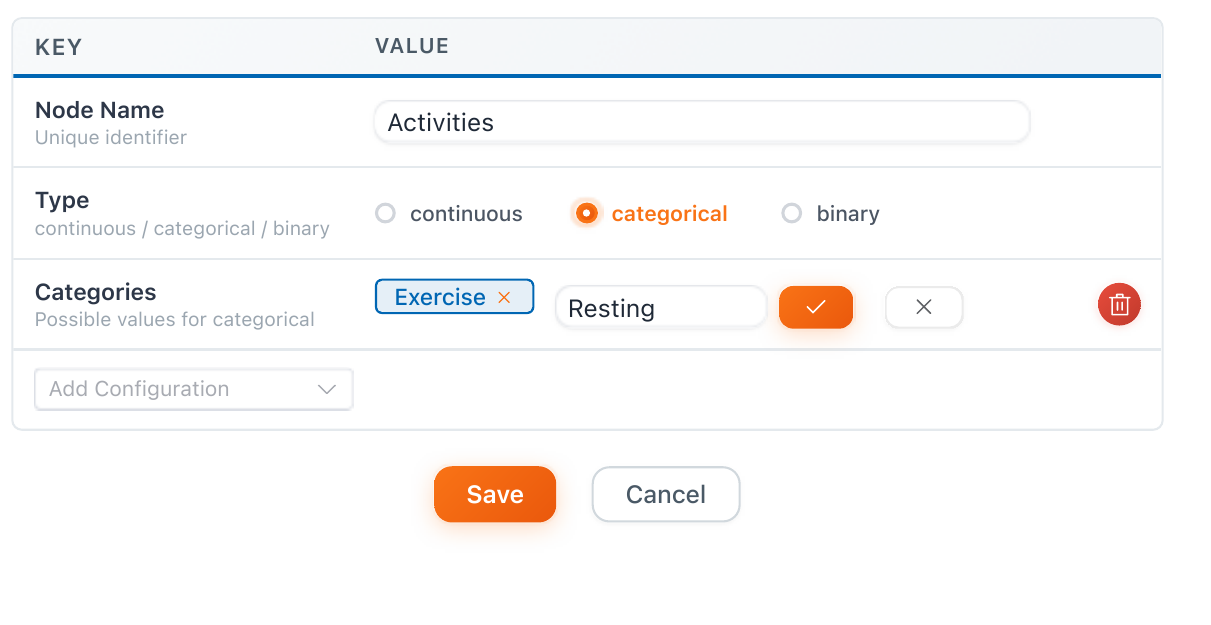}\label{fig:dialog:a}}
\subfigure[Node editor for continuous variables]{\includegraphics[width=0.99\linewidth, trim={10cm 1cm 1cm 26.2cm}, clip]{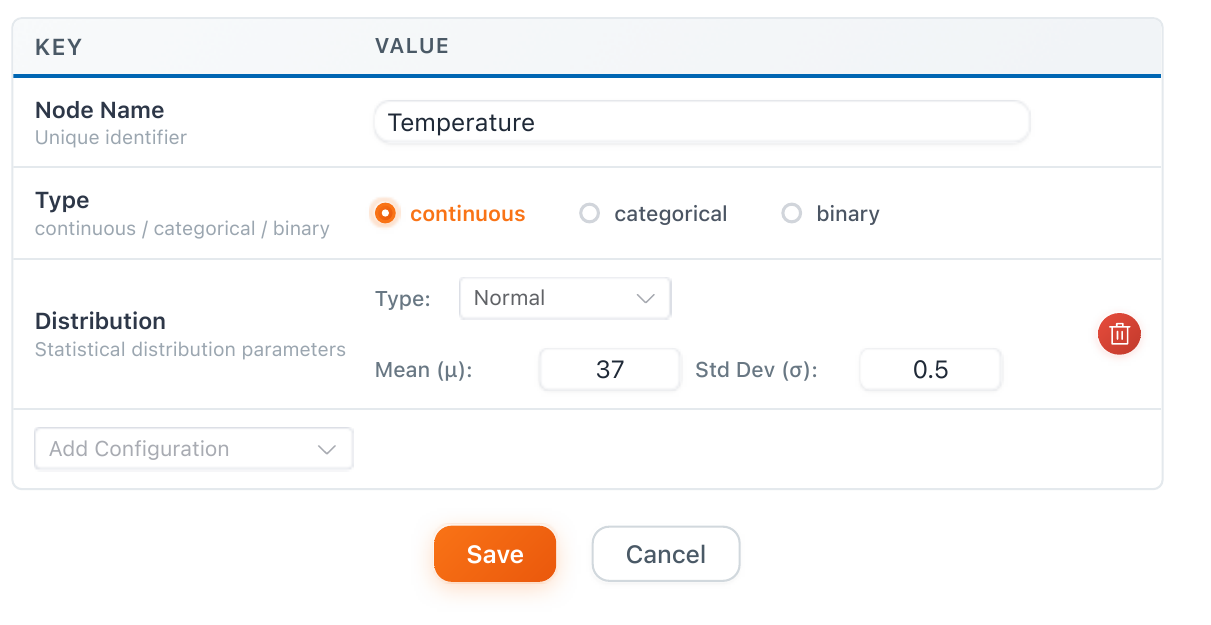}\label{fig:dialog:b}}
\caption{
Dialogues for node editing are illustrated using an example from a physiological scenario. 
}
\label{fig:dialog}
\end{figure}

\paragraph{Node Editing} Figure~\ref{fig:dialog:a} shows the input interface for adding or editing a categorical variable, such as “activities.” The user enumerates the possible values of interest by adding items; in the background, the categories are assigned unique integer labels so that the backend can utilize them seamlessly. Similarly, Figure~\ref{fig:dialog:b} shows an example of a continuous variable, “temperature.” The minimum and maximum fields define the range, and an offset records a typical value of the variable. Additional configurations include a memo field for notes to the algorithm developer, an aggregation mode (averaging, summing, voting, etc., depending on the variable type), and a typical range when the variable is continuous. 

\begin{figure}
    \centering
    \includegraphics[width=\linewidth, trim={10cm 1cm 1cm 25.2cm}, clip]{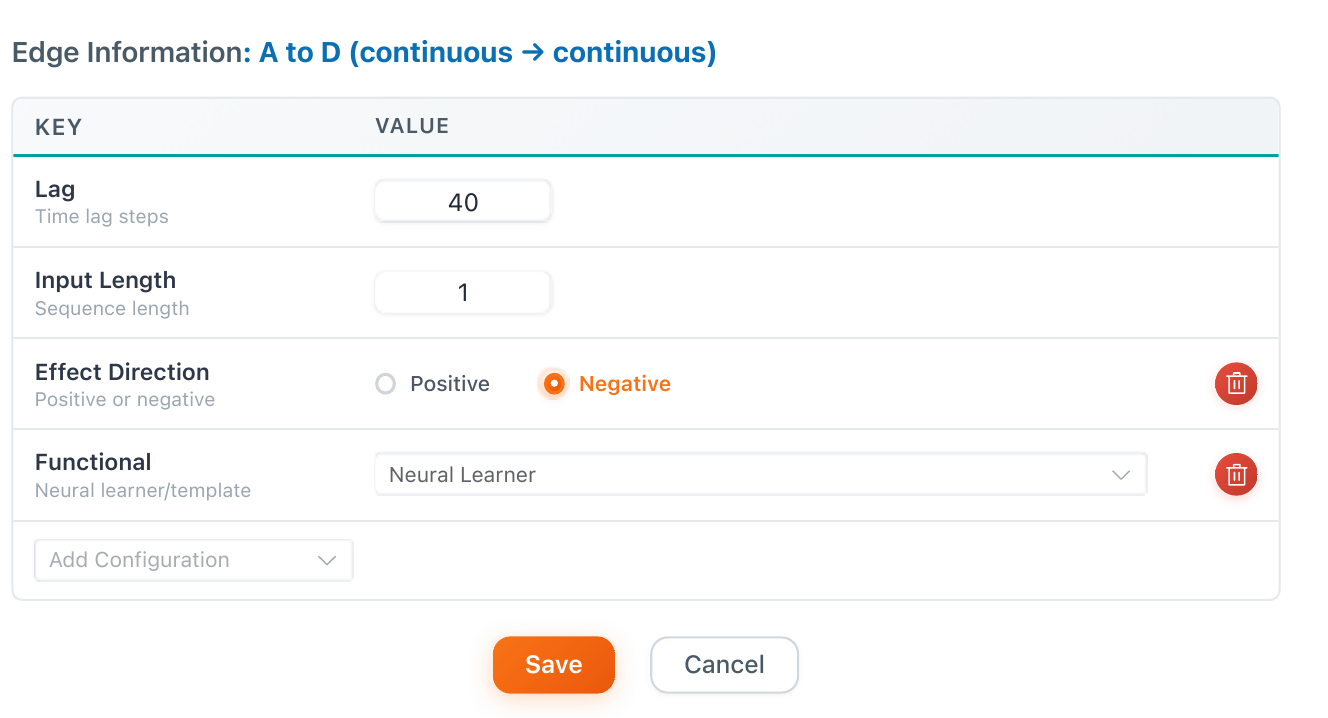}
    \caption{A dialogue for edge editing. }
    \label{fig:edge edit}
\end{figure}

\paragraph{Edge Editing}
Figure~\ref{fig:edge edit} illustrates how an edge can be edited. An important configuration is the choice of a functional, which can be either a statistical learner or a simple template, such as a linear model over a short window of the parent node’s history.

\section{Experiment Details}
\label{apdx:benchmark}

\subsection{Metrics}\label{sec:metrics}
\vspace{1px}
\paragraph{F1-score} We evaluate the performance of causal discovery algorithms by \textbf{F1 Score}, which provides a balanced measure of precision and recall, offering a comprehensive assessment of accuracy. The F1 Score is defined as:
$$
\text{F1-Score} = 2 \times \frac{\text{Precision} \times \text{Recall}}{\text{Precision} + \text{Recall}},
$$
where
$
\text{Precision} = \frac{|\text{TP}|}{|\text{TP}| + |\text{FP}|}
$
and
$
\text{Recall} = \frac{|\text{TP}|}{|\text{TP}| + |\text{FN}|}.
$
\vspace{1px}
In our experiment, true positives (\textbf{TP}) refer to cases where both the source and target nodes of an edge are correctly identified, and the lag falls within an acceptable window. False positives (\textbf{FP}) are incorrectly identified edges, while false negatives (\textbf{FN}) represent missing causal edges. 
For \emph{summary graphs}—in which edges are aggregated irrespective of their lags—an edge is deemed correctly identified if its source and target nodes correspond to those in the ground truth.

\paragraph{Structural Intervention Distance (SID)}
SID evaluates a learned causal graph, $\hat{G}$, by comparing it to a ground-truth graph, $G$. Unlike purely structural metrics like the Structural Hamming Distance (SHD), SID measures discrepancies in the causal effects implied by each graph.

Formally, SID is defined as the number of pairs of nodes $(i, j)$ for which the set of causal parents of $j$ under an intervention on $i$ differs between the two graphs. Let $\operatorname{Pa}_{G}(j | do(i))$ be the set of parents of node $j$ in the manipulated graph that results from an intervention on node $i$ in graph $G$. The SID is then calculated as:
$$ \text{SID}(\hat{G}, G) = \sum_{i \neq j} \mathbb{I} \left( \operatorname{Pa}_{\hat{G}}(j | do(i)) \neq \operatorname{Pa}_{G}(j | do(i)) \right) $$
where $\mathbb{I}(\cdot)$ is the indicator function. In simpler terms, a point is added to the distance for every pair of nodes where an intervention on one node leads to a different causal pathway to the other. This makes SID highly sensitive to errors in the causal hierarchy, providing a more robust measure of causal fidelity than metrics that simply count edge differences.
% For temporal graphs, correctness additionally requires the estimated lag to fall within a tolerance window around the true lag. To distinguish this tolerant evaluation from strict exact matching, we refer to metrics computed in this setting as \emph{relaxed}.

\subsection{Simulated Datasets}\label{sec:sim_data}

\begin{figure}[h]
\centering
\subfigure[]{\includegraphics[width=\linewidth]{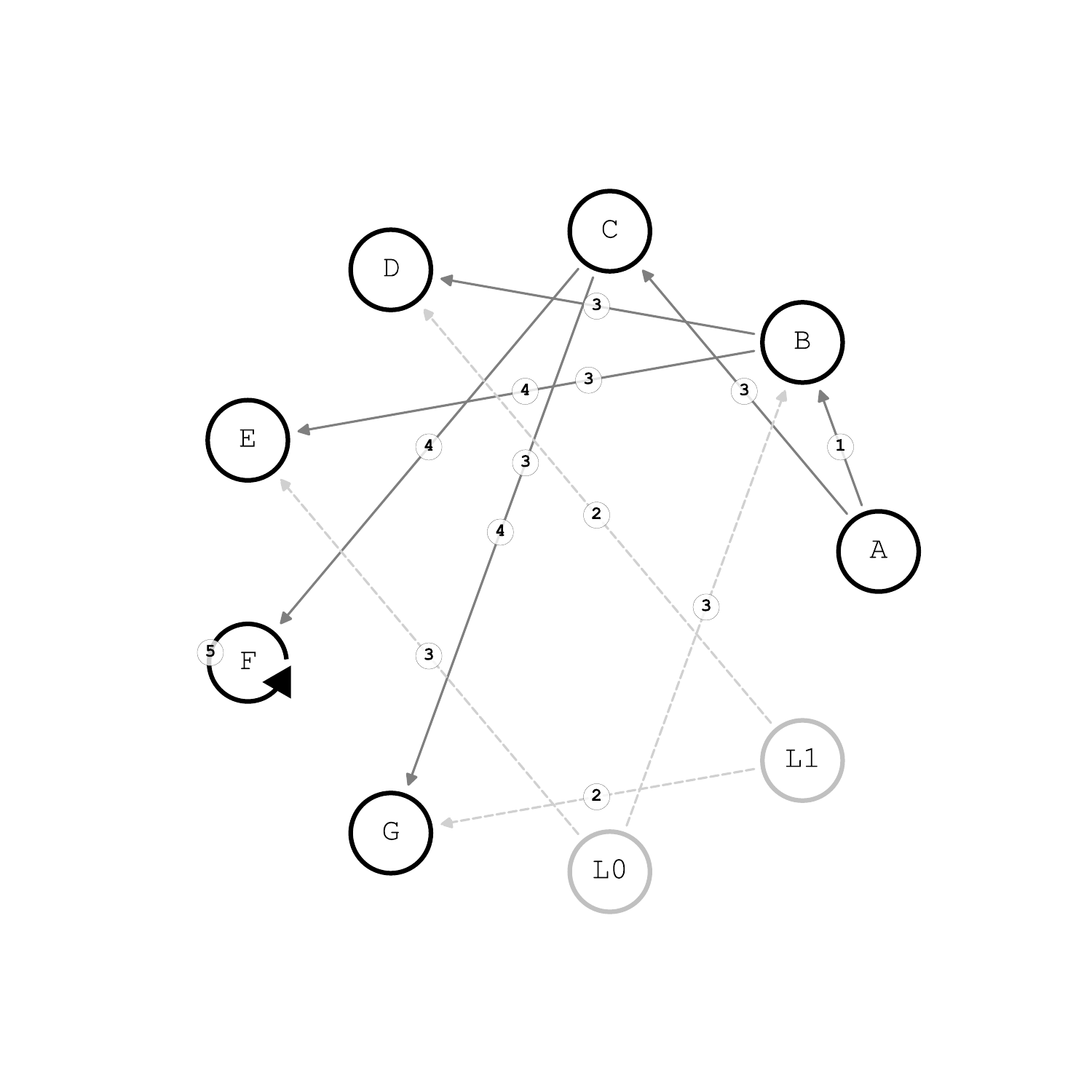}\label{fig:simdata:a}}
\subfigure[]{\includegraphics[width=\linewidth]{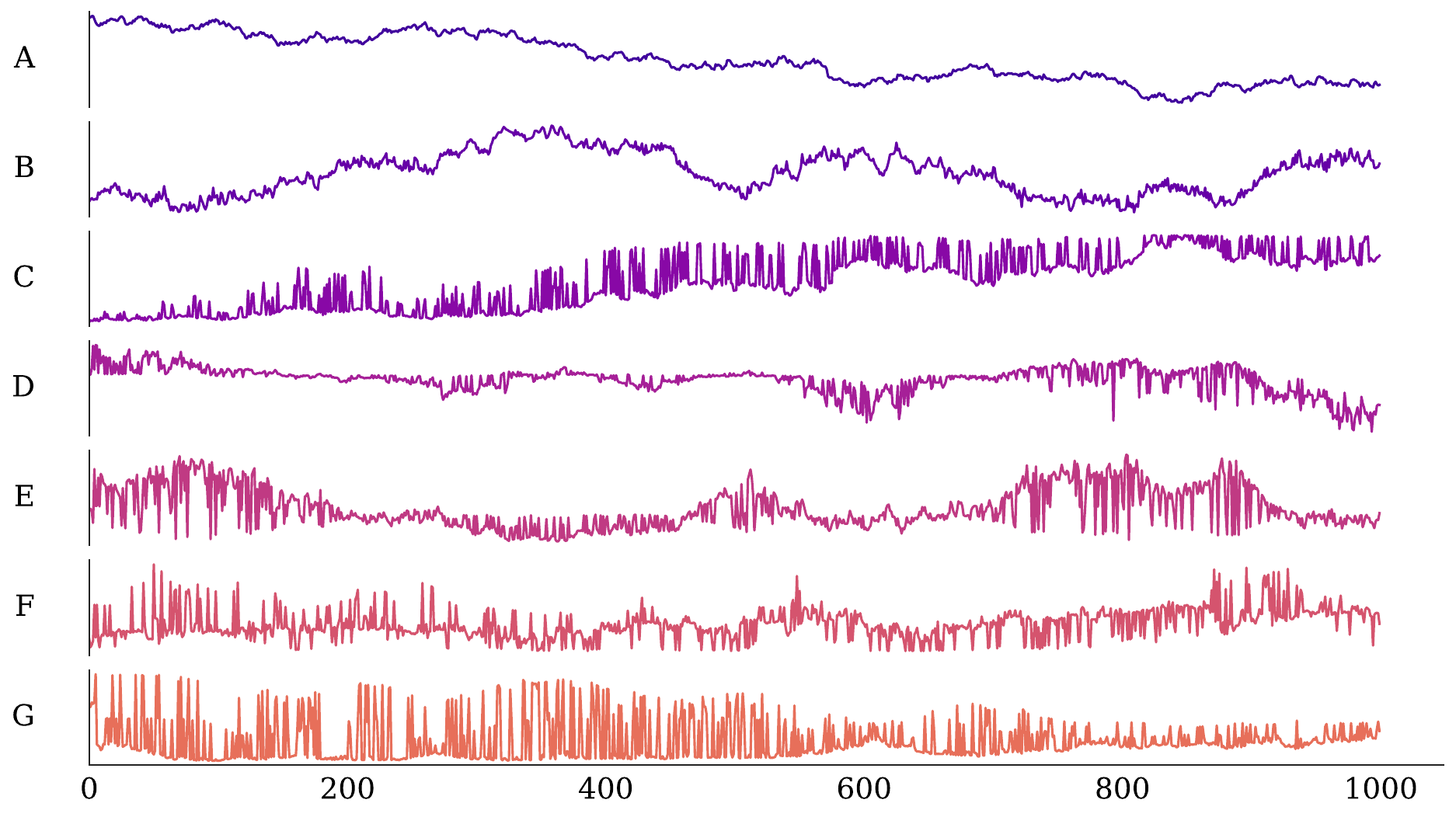}\label{fig:simdata:b}}
\caption{
Simulated data with latent variables.
}
\label{fig:simdata}
\end{figure}
We developed multiple simulated datasets featuring graph instances with a consistently fixed number of nodes of \textit{five}, allowing for an in-depth exploration of complex structural types while maintaining computational manageability. The ground-truth graphs in these datasets are template-based synthetic motifs (star, tree, cycle) with lagged edges, used as proxies for user- or expert-specified structures rather than graphs directly elicited from human experts; they are schema-compatible with true expert graphs in {KarmaTS} (same lagged/contemporaneous representation and export format), so real expert graphs can be imported and used interchangeably in the same benchmarking pipeline. The dataset is systematically varied along three major dimensions:  
\begin{enumerate}[wide, labelindent=0pt, label=\roman*)]
    \item \textit{Structural configurations.} We choose three classical graph topologies—\emph{star graphs}%
    % (the complete bipartite $K_{1,n}$)
    , \emph{cycle graphs}%
    % ($C_n$)
    , and \emph{(rooted) trees}—as defined in standard graph theory (e.g.\ ~\cite{harary1969graph}). They represent distinct challenges in causal inference and reflect common patterns observed in real-world networks.  

    \item \textit{Maximum lag of edges.} This parameter governs the temporal reach of causal links, allowing us to assess each algorithm’s ability to resolve delayed interactions inherent in dynamic systems. Here, we hold the edge–node ratio constant (see iii)). We evaluated two lag settings: (1) \textit{small}, with a maximum lag of 5 time units; and (2) \textit{large}, with a maximum lag of 10 time units.
    
    \item \textit{Edge–node ratio (ENR).} Defined as \(\lvert E\rvert / \lvert V\rvert\), the ENR approximates the average number of direct causes per node. By varying the ENR (including all lagged edges), we probe how algorithms perform under different connectivity and information densities. In this experiment, the maximum lag is fixed (see ii)). We consider two ENR regimes: (1) \emph{sparse}, with ENR \(\leq 2\); and (2) \emph{dense}, with \(2 < \mathrm{ENR} < 4\).

\end{enumerate}

We only considered lagged graphs in our simulation. This allows us to use cycles because a lag in edges prevents cyclicity. While it is possible to input contemporaneous effects for {KarmaTS}, we consider it a special case of a lag effect with a lower sampling frequency. For the functional maps, random multi-layer perceptrons (MLP) are employed.

We suggest that this paradigm could be enhanced by generating multivariate datasets with configurable spatio-temporal architectures. By allowing users to specify both the graphical model and the functional relationships between variables, we hope to enable the simulation of scenarios that better reflect real-world dynamics. For example, users might define spatial relationships, such as network-based or geographical dependencies, along with their temporal evolution patterns.

With the growing adoption of graphical model-based methodologies, there appears to be an increasing need for comprehensive benchmarking frameworks. Our proposed system aims to contribute to this space by providing a platform for evaluating time series forecasting and imputation algorithms using data that attempts to represent spatio-temporal complexities. Through the generation of data with adjustable characteristics, we hope that this work may help improve the assessment of algorithms across different scenarios, though further research would be needed to validate its effectiveness.

It is also possible to simulate datasets with latent variables. Figure~\ref{fig:simdata:a} shows a simulated graphical model representing the causal structure of seven \emph{observed} variables (A-G) and two \emph{latent variables} (L0 \& L1). The dashed arrows indicate the presence of a latent confounder influencing the system. No contemporaneous links are included in the model. Figure~\ref{fig:simdata:b} shows the corresponding simulated MTS, which was generated using a 2-layer multilayer perceptron (MLP) with ReLU activation functions.

\section{Factor-wise Analysis}\label{apdx:result}

\begin{table*}[h]
\renewcommand{\arraystretch}{1.2} % row spacing
\centering
\begin{tabular}{p{0.28\textwidth} p{0.68\textwidth}}
\toprule
\textbf{Benchmark / Setting} & \textbf{Reported strong methods (coarse ordering)} \\
\midrule
CAUSEME (stationary, moderate lag) &
\textbf{PCMCI}, \textbf{VAR}-based and \textbf{NOTEARS}-style methods near top; deep models more variable~\citep{rungeInferringCausationTime2019,pcmci,dynotears} \\
Synthetic irregular MTS (Neural ODE / \textbf{NGM}) &
\textbf{NGM} competitive in highly nonlinear continuous-time settings, but less stable on simpler graphs~\citep{ngm2021} \\
Attention-based \textbf{TCDF} studies &
\textbf{TCDF} strong on some nonlinear benchmarks but not consistently dominant across all structures~\citep{nauta2019causal} \\
High-dimensional graph benchmarks (\textbf{CUTS+}) &
\textbf{CUTS+} improves scalability but may sacrifice edge-wise accuracy in dense or long-lag regimes~\citep{cheng2024cuts+} \\
\midrule
{KarmaTS} summary (this work) &
\textbf{PCMCI} and \textbf{DYNOTEARS} typically dominate F1/SID; \textbf{TCDF} excels in some star-like structures; \textbf{NGM} and \textbf{CUTS+} lag on most settings but remain competitive in selected regimes (e.g., sparse or highly nonlinear) \\
\bottomrule
\end{tabular}
\caption{Coarse external calibration of method rankings from {KarmaTS} to published time-series causal discovery benchmarks.}
\label{tab:external-calibration}
\end{table*}

\begin{figure}[ht]
    \centering
    \includegraphics[width=\linewidth]{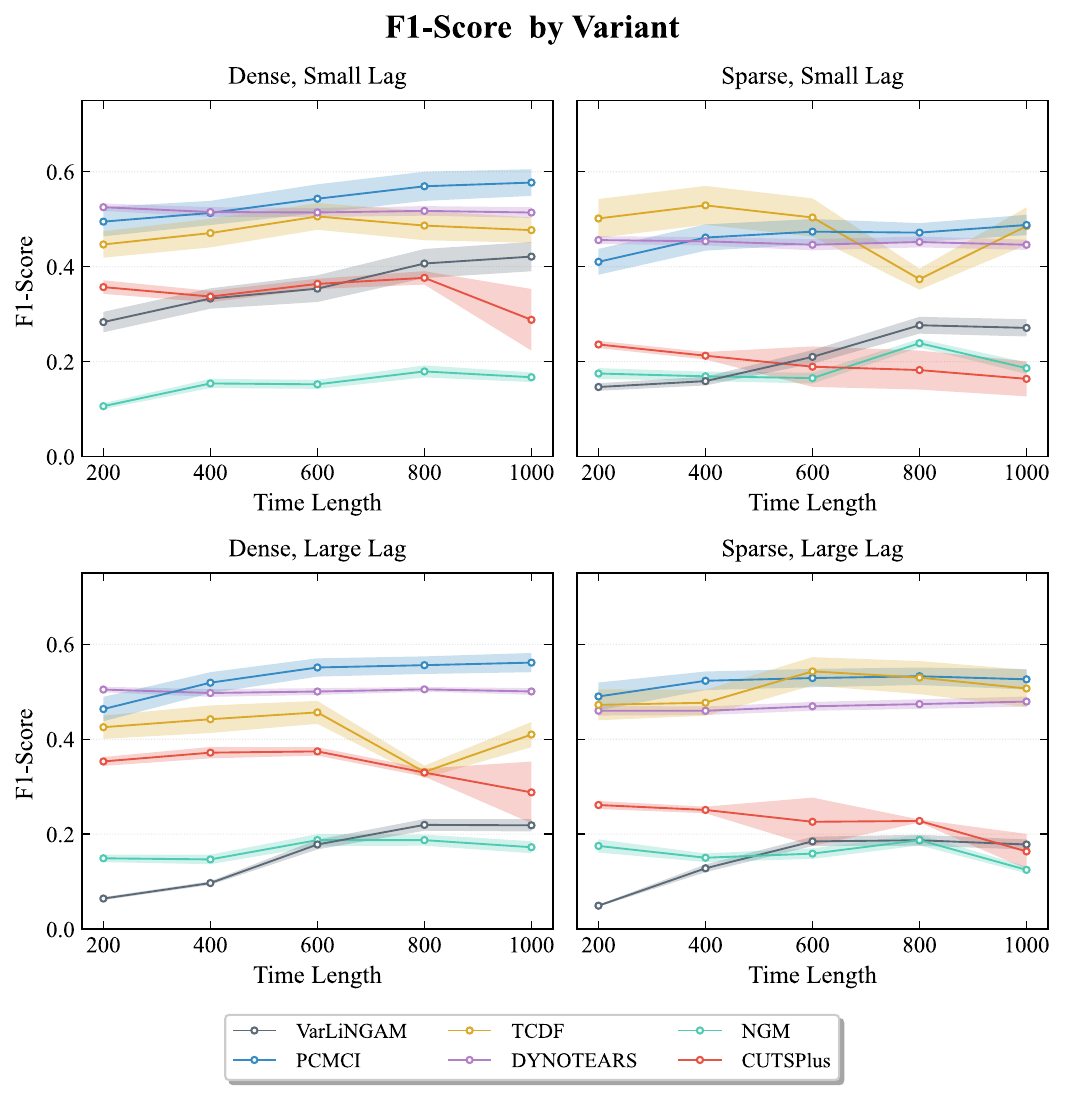}
    \caption{F1-score performance grouped by ENR and maximum lag of edges; results are averaged over all structural configurations (see Section~\ref{sec:sim_data}). }
    \label{fig:fourplot}
\end{figure}
\paragraph{Time Series Length}
In the main text, Figure~\ref{fig:exp-samplesize} shows F1 versus sequence length (\(100\!-\!1000\)). Constraint-based methods (\textbf{PCMCI}, \textbf{VarLiNGAM}) benefit most from longer series, gaining up to \(+0.25\) F1, whereas gradient-based (\textbf{DYNOTEARS}) plateaus early once its acyclicity penalty stabilizes.

\begin{figure}[ht]
    \centering
    \includegraphics[width=\linewidth]{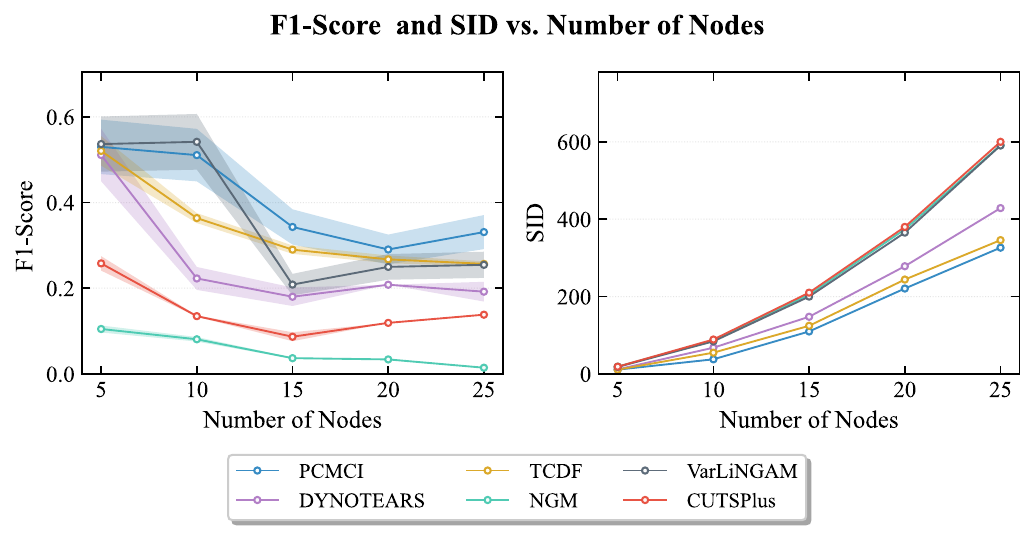}
    \caption{
    F1-score and SID for causal discovery methods, as a function of \textbf{the number of nodes} at a fixed time series length of 600. Formatting conventions are detailed in Section~\ref{misc:plot}.
    }
    \label{fig:n_nodes_metrics}
\end{figure}

\paragraph{Edge Density and Maximum Lag}
Model performance also varies significantly with data density and time lag, as shown in Figure~\ref{fig:fourplot}. For dense data, \textbf{PCMCI} consistently holds an advantage, achieving the highest F1-Score of $\sim$0.6 with a small lag and $\sim$0.55 with a large lag. In sparse conditions, however, the best-performing model changes. With a small lag, \textbf{TCDF} demonstrates a clear advantage with a peak F1-score of $\sim$0.55. With a large lag under sparse conditions, both \textbf{PCMCI} and \textbf{TCDF} emerge as the most robust choices. This again highlights that performance is not universal, but rather a function of specific data characteristics.

\paragraph{Number of Nodes}
The result is shown in Figure~\ref{fig:n_nodes_metrics}. As the graph size increases, F1 Scores decline and SID rises precipitously. \textbf{PCMCI} preserves the highest F1 on smaller graphs, while \textbf{CUTS+} and \textbf{TCDF} exhibit comparatively greater robustness in SID.

\begin{figure}[ht]
    \centering
    \begin{center}
       {\includegraphics[width=\linewidth]{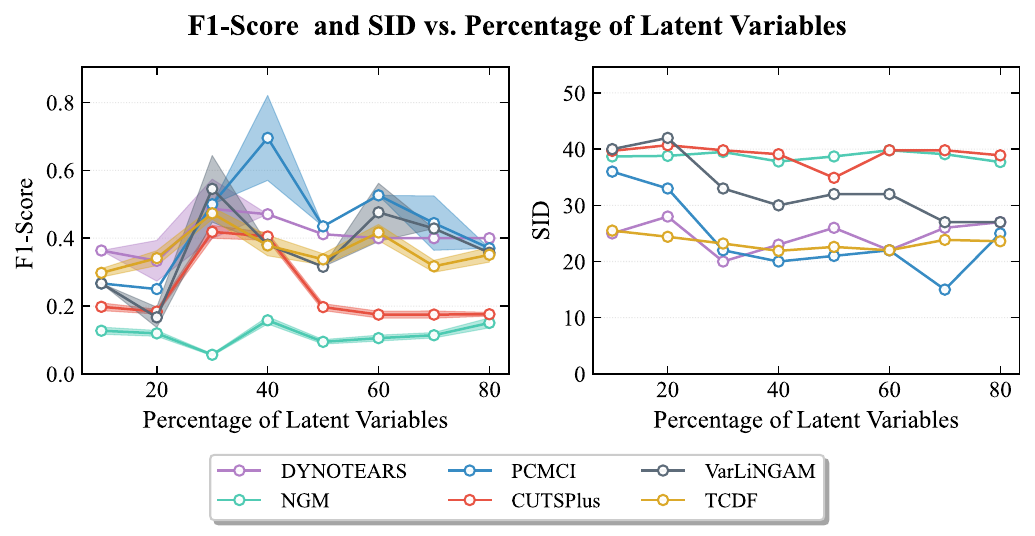}}
    \end{center}
    \caption{
    F1-score and SID for causal discovery methods. The time series length is fixed to be 600. Plot formatting details are given in Section~\ref{misc:plot}.
    }
    \label{fig:exp-latent}
\end{figure}

\paragraph{Latent Variables}
Figure~\ref{fig:exp-latent} plots F1 as the latent-variable fraction rises from \(0\%\) to \(80\%\). The figure shows that the effect of latent variables on performance does not follow a simple monotonic trend: both F1-score and SID fluctuate as the percentage of latent variables increases. This behavior is consistent with prior findings that latent variables do not necessarily degrade or improve causal discovery, but can either obscure or reveal structure depending on the setting and assumptions \citep{LPCMCI, dong2024rlcd, cai2023cumulants}.

\section{External Calibration to Prior Benchmarks}
\label{apdx:external_calibration}

Table~\ref{tab:external-calibration} provides a coarse alignment between the relative performance we observe on {KarmaTS} and method orderings reported in prior time-series causal discovery benchmarks, such as CAUSEME and follow-up studies~\citep{rungeInferringCausationTime2019,gong2024causal,pcmci,dynotears,nauta2019causal,ngm2021,cheng2024cuts+}. The goal is not to re-benchmark these methods, but to indicate which patterns appear consistent across settings versus those that appear specific to our stress tests.

\end{document}